\documentclass[a4paper]{cas-sc}

\usepackage[authoryear,longnamesfirst]{natbib}
\usepackage{textcomp}
\usepackage[protrusion=true,expansion=false]{microtype}
\usepackage{pdfpages}
\usepackage{cleveref}
\usepackage{tabularx}
\usepackage{enumitem}
\usepackage{pifont}
\usepackage{wasysym}
\usepackage{tikz}
\usetikzlibrary{arrows.meta,positioning,shapes.geometric,calc}
\usepackage{pgfplots}
\pgfplotsset{compat=1.18}
\usepgfplotslibrary{fillbetween}
\PassOptionsToPackage{table}{xcolor}

\hypersetup{
  colorlinks=true,
  linkcolor=blue!60!black,
  citecolor=blue!60!black,
  urlcolor=blue!60!black
}

\newcommand{\ecm}{\textsc{ECM}}
\newcommand{\candidate}{\hat{c}}
\newcommand{\policyset}{\Pi}
\newcommand{\upgfunc}{\mathcal{U}}

\definecolor{orcidlogocol}{HTML}{A6CE39}
\newcommand{\orcidicon}{%
  \mbox{\begin{tikzpicture}[x=1ex,y=1ex,baseline={([yshift=-0.45ex]current bounding box.center)}]%
    \fill[orcidlogocol] (0,0) circle (0.95);%
    \node[white,font=\fontsize{3pt}{3pt}\selectfont\bfseries\sffamily] at (0,0) {iD};%
  \end{tikzpicture}}%
}
\newcommand{\orcidlink}[1]{\href{https://orcid.org/#1}{\orcidicon}}

\definecolor{paperblue}{HTML}{1F6DC7}   
\definecolor{paperbluel}{HTML}{BDD4EE}  
\definecolor{paperred}{HTML}{C0392B}    
\definecolor{papergray}{HTML}{6B7280}   
\definecolor{paperrowhi}{HTML}{E8F1FB}  

\newtheorem{definition}{Definition}[section]

\usepackage{lastpage}
\AtBeginDocument{%
  \makeatletter
  \def\lastpage{\pageref{LastPage}}%
  \makeatother
}

\begin{document}

\let\WriteBookmarks\relax
\def\floatpagepagefraction{1}
\def\textpagefraction{.001}

\shorttitle{Governed Capability Evolution}
\shortauthors{X. Qin et~al.}

\title[mode=title]{Governed Capability Evolution: Lifecycle-Time Compatibility Checking and Rollback for AI-Component-Based Systems, with a Proof-of-Concept Evaluation on Embodied Agents}

\author[1]{Xue Qin}[orcid=0009-0009-3642-2663]
\ead{qinxue@me.com}
\credit{Conceptualization, Methodology, Software, Investigation, Writing -- original draft}

\author[2]{Simin Luan}[orcid=0000-0003-1138-1892]
\ead{luansiminiot@gmail.com}
\credit{Methodology, Validation}

\author[3]{John See}[orcid=0000-0003-3005-4109]
\ead{J.See@hw.ac.uk}
\credit{Writing -- review \& editing}

\author[5]{Zeyd Boukhers}[orcid=0000-0001-9778-9164]
\ead{zeyd.boukhers@fit.fraunhofer.de}
\credit{Writing -- review \& editing}

\author[4]{Cong Yang}[orcid=0000-0002-8314-0935]
\cormark[1]
\ead{cong.yang@suda.edu.cn}
\credit{Supervision, Writing -- review \& editing}

\author[2]{Zhijun Li}[orcid=0000-0001-9129-9957]
\cormark[1]
\ead{lizhijun_os@hit.edu.cn}
\credit{Supervision, Writing -- review \& editing}

\affiliation[1]{organization={School of Software, Harbin Institute of Technology},
                city={Harbin},
                country={China}}
\affiliation[2]{organization={School of Computer Science and Technology, Harbin Institute of Technology},
                city={Harbin},
                country={China}}
\affiliation[3]{organization={School of Mathematical and Computer Sciences, Heriot-Watt University, Malaysia Campus},
                city={Putrajaya},
                country={Malaysia}}
\affiliation[4]{organization={School of Future Science and Engineering, Soochow University},
                city={Suzhou},
                country={China}}
\affiliation[5]{organization={Fraunhofer Institute for Applied Information Technology},
                city={Sankt Augustin},
                country={Germany}}

\cortext[cor1]{Corresponding authors}
\nonumnote{\copyright\ 2026. This manuscript version is made available under the CC-BY-NC-ND 4.0 license \url{https://creativecommons.org/licenses/by-nc-nd/4.0/}. It is the accepted manuscript of an article published in \emph{Information and Software Technology} 200 (2026) 108337, \url{https://doi.org/10.1016/j.infsof.2026.108337}. The Supplementary Material is appended after the references.}

\begin{abstract}
\textbf{Context.} Software systems built from versioned AI components need lifecycle-time governance over how a new module version is admitted, monitored, and withdrawn. Established deployment patterns (canary, blue-green, feature flags, MLOps) monitor aggregate service-level signals of stateless services, not the stateful, policy-constrained runtimes that drive AI components. \textbf{Objectives.} We formulate \emph{governed capability evolution} as a software-lifecycle problem, and ask whether staged upgrade governance prevents unsafe activations where practice-adapted canary and blue-green deployment do not, and at what cost. \textbf{Methods.} We propose a staged upgrade framework with four compatibility checks (interface, policy, behavioral, recovery) and a seven-stage pipeline (validation, sandbox, shadow, gated activation, monitoring, rollback, audit). A reference prototype is evaluated in a behavioral simulation of an embodied manipulation stack, with a PyBullet backend replicating the strategy ordering. Upgrade candidates come from a seeded fault-operator generator (45 per capability family and seed, including out-of-taxonomy probes) rather than hand-picking, and thresholds are fixed on a disjoint development set. We compare five strategies over five upgrade rounds and 15 seeds with paired Wilcoxon tests under Holm--Bonferroni correction. \textbf{Results.} Naive upgrade reaches 73.3\% final-round success but activates unsafe versions in 74.7\% of rounds. Canary and blue-green adaptations keep comparable success yet still admit faulty versions in 61.3\% and 64.0\% of upgrade rounds, because aggregate success hides policy and recovery regressions. Governed upgrade admits no in-taxonomy fault and one out-of-taxonomy probe in 75 behavioral-track rounds, at a 4.6-point success cost (all contrasts against the three baselines: adjusted $p\!\le\!0.024$); the full pre-activation stack detects 97.3--100\% of in-taxonomy faults across deployment profiles, and rollback restores verified operation in 72.4\% of drift scenarios. \textbf{Conclusion.} Bounded-window aggregate-success gating proved insufficient against these fault operators; staged governance closes the gap at a modest, tunable cost on existing CI/CD infrastructure.
\end{abstract}


\maketitle
\hypersetup{pdftitle={Governed Capability Evolution: Lifecycle-Time Compatibility Checking and Rollback for AI-Component-Based Systems},pdfauthor={Xue Qin, Simin Luan, John See, Zeyd Boukhers, Cong Yang, Zhijun Li}}

\section{Introduction}
\label{sec:introduction}

Embodied agents are increasingly expected to operate as long-lived systems that persist across tasks, environments, and deployment phases, with capabilities added, improved, or replaced over time. Capability upgrade is therefore a normal operating condition rather than an exceptional event. The same shift runs through AI-component-based software more broadly: perception models, skill policies, and planning components ship as versioned artifacts that a deployed system is expected to absorb without stopping. Each upgrade can change interfaces, alter behavior distributions, expand permission requirements, or break recovery assumptions. It does so inside a runtime that carries state and operates under deployment policy.

\begin{figure}[pos=t!p]
\centering
\includegraphics[width=0.6\linewidth]{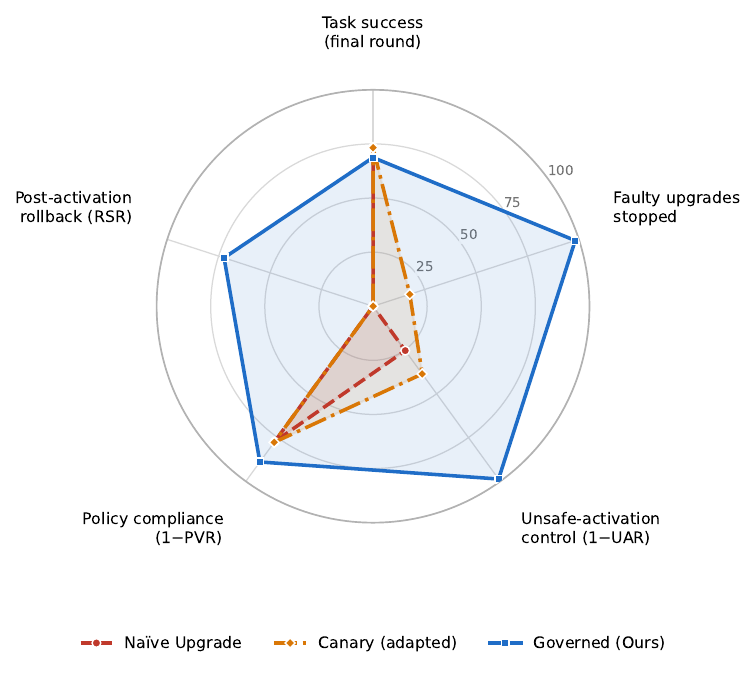}
\caption{Governance profile comparison across five deployment dimensions (all axes oriented outer\,=\,better; values from the 15-seed evaluation of \Cref{sec:results}). Governed Upgrade (blue) covers the safety and recoverability dimensions while staying competitive on task success. The practice-adapted Canary baseline (orange) improves on Na\"ive only marginally: it aborts the 17.9\% of faulty upgrades whose cohort success rate visibly collapses and promotes the rest. Na\"ive Upgrade (red) collapses on four of the five dimensions, retaining only partial policy compliance.}
\label{fig:radar-teaser}
\end{figure}

How are such upgrades shipped today? The state of the practice comes from service deployment. Canary releases route a small share of traffic to the new version and compare aggregate health signals against the incumbent before promotion~\citep{canary,stagedrollout,autorollout}; blue-green deployment keeps the previous version warm and cuts over with a fast revert path~\citep{canary}; feature flags and progressive delivery stage exposure across cohorts~\citep{progressivedelivery}; MLOps pipelines add offline validation and staged rollout for learned models~\citep{paleyes2022challenges,ashmore2021assuring,sculley2015techdebt}. These patterns work because the signal they monitor (error rate, latency, aggregate success) is a faithful proxy for what can go wrong in a stateless service. For the stateful, policy-constrained runtimes that drive AI components, the proxy breaks (\Cref{fig:radar-teaser}). A nominally stronger version can raise task success while violating deployment policy more often. It can widen its permission footprint with no visible service degradation, and weaken the recovery paths the system depends on exactly when something else fails. Aggregate service health sees none of this until the damage is live.

This paper formulates the resulting problem as \emph{governed capability evolution}: a newly produced capability version is a governed deployment candidate whose admission must itself be evaluated. This extends runtime governance from the execution lifecycle to the capability-lifecycle boundary. Instead of asking only whether an action invocation should be allowed, we ask whether a new version should be admitted, under what conditions it may be activated, how it should be monitored, and when it should be rolled back. We introduce four upgrade-oriented compatibility dimensions (interface, policy, behavioral, recovery) and organize them into a staged upgrade pipeline from isolated evaluation to monitored live deployment, with rollback available at every stage. The prototype builds on a single-agent embodied runtime with modular capability packaging and execution-time governance~\citep{aeros,identity,harnessing}; the present paper adds the lifecycle layer that governs how new capability versions enter that running system. \Cref{fig:overview} contrasts the na\"ive and governed paths.

\begin{figure*}[pos=t!p]
\centering
\begin{tikzpicture}[
    >=stealth,
    scale=0.88, every node/.style={scale=0.88},
    stageBase/.style={rectangle, rounded corners=3pt,
                      minimum height=0.72cm, minimum width=1.55cm, align=center,
                      font=\scriptsize\sffamily, line width=0.6pt},
    stageLight/.style={stageBase, draw=paperblue, fill=paperbluel},
    stageMed/.style={stageBase,   draw=paperblue, fill=paperblue!30},
    stageDeep/.style={stageBase,  draw=paperblue!85!black, fill=paperblue, text=white},
    stageFail/.style={stageBase,  draw=paperred,  fill=paperred!14},
    stageNeutral/.style={stageBase, draw=papergray, fill=gray!6},
    fadenode/.style={stageBase, draw=papergray!70, fill=gray!8, text=papergray},
    arrowBlue/.style={->, thick, color=paperblue},
    arrowRed/.style={->, thick, color=paperred},
    arrowGray/.style={->, thick, color=papergray},
    arrowRedDash/.style={->, dashed, thick, color=paperred},
    arrowGrayDash/.style={->, dashed, thick, color=papergray!75},
    arrowBlueDot/.style={->, densely dotted, thick, color=paperblue},
]

\node[font=\scriptsize\sffamily\bfseries, color=paperred, anchor=east]
  at (-0.5, 2.8) {Na\"ive Upgrade:};

\node[stageNeutral] (n-learn)   at (1.0, 2.8) {Capability\\Learning};
\node[stageFail]    (n-replace) at (3.8, 2.8) {Immediate\\Replace};
\node[stageFail]    (n-run)     at (6.6, 2.8) {Run in\\Production};
\node[fadenode]     (n-hope)    at (9.4, 2.8) {Hope for\\the Best};

\draw[arrowGray]    (n-learn)   -- (n-replace);
\draw[arrowRed]     (n-replace) -- (n-run);
\draw[arrowRedDash] (n-run)     -- (n-hope);

\node[font=\scriptsize\sffamily, color=paperred, align=center] at (6.6, 2.1)
  {no compatibility check,\; no monitoring,\; no rollback};

\draw[gray!40, dashed] (-1.5, 1.5) -- (14.2, 1.5);

\node[font=\scriptsize\sffamily\bfseries, color=paperblue, anchor=east]
  at (-0.5, 0.5) {Governed Upgrade:};

\node[stageLight] (g-reg) at (1.0, 0.5)  {Candidate\\Register};
\node[stageLight] (g-val) at (3.4, 0.5)  {Compat.\\Validate};
\node[stageMed]   (g-sb)  at (5.8, 0.5)  {Sandbox\\Evaluate};
\node[stageMed]   (g-sh)  at (8.2, 0.5)  {Shadow\\Deploy};
\node[stageDeep]  (g-act) at (10.6, 0.5) {Gated\\Activate};
\node[stageDeep]  (g-mon) at (13.0, 0.5) {Online\\Monitor};

\draw[arrowBlue] (g-reg) -- (g-val);
\draw[arrowBlue] (g-val) -- (g-sb);
\draw[arrowBlue] (g-sb)  -- (g-sh);
\draw[arrowBlue] (g-sh)  -- (g-act);
\draw[arrowBlue] (g-act) -- (g-mon);

\node[stageFail]    (g-rej) at (3.4, -1.2)  {Reject};
\node[stageNeutral] (g-dem) at (8.2, -1.2)  {Demote};
\node[stageFail]    (g-rb)  at (13.0, -1.2) {Rollback};

\draw[arrowRedDash]  (g-val) -- node[right, font=\scriptsize\sffamily, text=paperred] {fail} (g-rej);
\draw[arrowGrayDash] (g-sb.south) -- ++(0,-0.3) -| (g-dem);
\draw[arrowGrayDash] (g-sh.south) -- (g-dem);
\draw[arrowRedDash]  (g-mon) -- node[right, font=\scriptsize\sffamily, text=paperred] {drift} (g-rb);

\draw[arrowBlueDot] (g-rb.south) -- ++(0,-0.4) -|
  node[pos=0.25, below, font=\scriptsize\sffamily, text=paperblue]
  {restore previous active version} (g-act.south);

\node[font=\scriptsize\sffamily, color=paperblue] at (3.4, 1.15)  {$\kappa_I,\kappa_P,\kappa_B,\kappa_R,\ \theta_{\mathrm{act}}$};
\node[font=\scriptsize\sffamily, color=paperblue] at (5.8, 1.15)  {$\theta_{\mathrm{sb}}$};
\node[font=\scriptsize\sffamily, color=paperblue] at (8.2, 1.15)  {$\Delta_{\mathrm{shadow}}$};
\node[font=\scriptsize\sffamily, color=paperblue] at (10.6, 1.15) {approval};

\end{tikzpicture}
\caption{Overview of governed capability evolution. \textbf{Top:} Na\"ive upgrade directly replaces the active capability version without governance. \textbf{Bottom:} Each new version is a candidate that must pass compatibility validation, which scores all four dimensions ($\kappa_I, \kappa_P, \kappa_B, \kappa_R$) from declared metadata before any execution, against the composite gate $\theta_{\mathrm{act}}$, then the sandbox envelope $\theta_{\mathrm{sb}}$, the shadow divergence bound $\Delta_{\mathrm{shadow}}$, and gated activation, which adds supervisory approval in approval-requiring profiles. Online monitoring continues after activation; \Cref{fig:fsm} gives the full transition set, including demotion and rollback.}
\label{fig:overview}
\end{figure*}

We evaluate a reference prototype in a behavioral simulation of an embodied manipulation stack, with a PyBullet physics backend replicating the headline comparison. Upgrade candidates are not hand-picked: a seeded generator samples benign and faulty candidates from parameterized fault-operator families, and adds out-of-taxonomy probes that none of the four checks models, so the evaluation also measures where the pipeline's coverage ends. Five strategies are compared over repeated upgrade rounds: static (never upgrade), na\"ive replacement, practice-adapted canary and blue-green deployment, and the governed pipeline.

This paper makes four contributions.
\begin{itemize}[leftmargin=*,nosep]
    \item We formulate governed capability evolution as a software-lifecycle problem and define a four-dimensional upgrade compatibility model (interface, policy, behavioral, recovery) over declared capability metadata.
    \item We design a seven-stage governed upgrade pipeline with explicit state-machine semantics, from candidate validation through gated activation to rollback and audit.
    \item We release a reference implementation and an operator-based evaluation methodology: faulty candidates are sampled from parameterized fault families disjoint from the threshold development set, and out-of-taxonomy probes measure coverage boundaries.
    \item We show empirically that practice-adapted canary and blue-green deployment admit faulty versions in 61.3\% and 64.0\% of upgrade rounds respectively, while the governed pipeline admits no in-taxonomy fault, and one out-of-taxonomy probe, in 75 behavioral-track rounds at a 4.6-point success cost (15 seeds, paired Wilcoxon tests, Holm--Bonferroni corrected).
\end{itemize}

\Cref{sec:related-work} reviews related work; \Cref{sec:problem,sec:compatibility,sec:pipeline} develop the problem formulation, the compatibility model, and the pipeline; \Cref{sec:implementation,sec:experiments,sec:results} cover the prototype, the experimental setup, and the results with ablations; \Cref{sec:discussion,sec:threats,sec:conclusion} close with discussion, threats to validity, and conclusions.

\section{Related Work}
\label{sec:related-work}

Four bodies of prior work bear on governed capability evolution: robotic architectures and capability organization, modular skill learning, safe robotics and runtime governance, and runtime enforcement for LLM agents. We review each in turn and then relate our design to software deployment practice. Each area supplies ingredients; none governs how a new capability version enters a running embodied system.

\paragraph{Robotic architectures, middleware, and modular skills}
Middleware research established components as the unit of robot software: ROS and ROS~2 organize a robot as a graph of nodes with typed interfaces~\citep{ros,ros2}, and comparable models underpin OROCOS and YARP~\citep{componentbased,yarp}. ROS~2's managed lifecycle nodes go furthest toward our concern: each node carries an explicit state machine over configuring, activating, and deactivating~\citep{ros2}. That machinery governs the activation state of a running node; it says nothing about whether a \emph{new version} of the node should be trusted, so version admission stays with the human operator. Skill-level abstractions raise the unit of composition from node to capability. SkiROS attaches pre- and post-conditions to skills inside a shared world model~\citep{skiros}, and behavior trees compose such skills reactively~\citep{behaviortrees}; both assume the skill set is curated offline before deployment. LLM-driven stacks sharpen the problem: Code as Policies generates fresh capability code at runtime~\citep{codepolicies}, Inner Monologue re-plans around feedback~\citep{innermonologue}, and foundation-model controllers such as RT-2 fold many skills into one large policy~\citep{rt2}. Newly generated or newly fine-tuned behavior thus typically enters the system without an admission gate. AEROS packages capabilities as installable modules with declared interfaces and permissions under a policy-separated runtime~\citep{aeros}; it defines where an admission gate could sit, without defining the gate.

Research on modular skill learning explains where the new versions come from. Hierarchical RL extracts reusable options~\citep{options}; SPiRL and SkiMo learn skill priors and skill dynamics for downstream reuse~\citep{spirl,skimo}; continual-learning agents accumulate skills over a lifetime~\citep{lifelong,continual}. Voyager is the sharpest instance: it writes, refines, and stores new skill code continuously as it plays~\citep{voyager}. PhyAgentOS pushes the same trajectory to the operating-system level, self-evolving an embodied agent's capability library with decoupled cognitive planning and physical execution~\citep{liu2026phyagentos}; its pipeline contains no compatibility checking, no staged admission, and no rollback governance, which is precisely the lifecycle layer studied here. Capability-centric evolution~\citep{identity} adds version registries and gated deployment for learned modules, and stops short of making the admission decision itself the object of study.

A software-engineering view of this space is consolidating. Skillware treats agent skills as software artifacts with a managed lifecycle including removal and rollback~\citep{skillware2026}; a recent survey taxonomizes evolving skill libraries with explicit verification and governance stages~\citep{li2026skilllifecycle}; falsifiable release gates hold standing safety invariants across releases of self-improving systems~\citep{soni2026releasegates}; and anytime-valid certificates gate self-modification against statistical error budgets~\citep{sengupta2026certificates}. These concurrent proposals confirm the lifecycle framing and sharpen what remains open: none couples admission to a runtime governance layer with staged dynamic evaluation, monitored activation, and rollback for the stateful, policy-constrained runtimes studied here.

\paragraph{Safe robotics, runtime governance, and LLM-agent guardrails}
Runtime safety research supplies the enforcement vocabulary this paper builds on: control barrier functions keep the state inside a safe set~\citep{cbf}, Simplex switches to a verified fallback controller when the advanced one misbehaves~\citep{simplex}, shielding filters unsafe actions from a learned policy's output~\citep{shielding}, and safe-RL methods move the constraint into training~\citep{saferl,brunke2022safe}. Runtime monitors, assurance architectures, and formal verification of autonomous systems detect or prove properties on the deployed executable~\citep{runtimeverification,hobbs2023runtime,luckcuck2019formal,seshia2022verified}, and K{\"o}nighofer et al.\ organize the field by distinguishing runtime enforcement (block, edit, replay) from runtime verification (post-hoc detection)~\citep{konighofer2022re}. Every one of these mechanisms presumes a \emph{fixed} executable system: the controller being shielded today is assumed to be the controller that was certified yesterday. Capability upgrade breaks that assumption. Harnessing Embodied Agents reframes enforcement as a systems layer covering admission, policy checks, execution watching, recovery, and audit~\citep{harnessing}; we extend that layer from capability invocation to capability replacement, from governing what runs to governing what is admitted to run.

LLM-agent safety develops the same enforcement ideas for language-driven agents: AutoRT screens task proposals against a constitution~\citep{autort}, RoboGuard grounds safety rules into temporal-logic plan checks~\citep{roboguard}, SafeEmbodAI hardens prompt and perception channels~\citep{safeembodai}, AgentSpec and NeMo Guardrails let operators write enforceable action rules~\citep{agentspec,nemoguardrails}, and TrustAgent and ProbGuard add stage-wise and probabilistic intervention~\citep{trustagent,pro2guard,swisscheese}. These mechanisms gate the plans or actions of one agent version; the capability that produced the plan is not itself versioned or re-admitted when it changes. Two proposals come closest to our framing. MI9~\citep{mi9} is a runtime governance framework for agentic AI with six execution-time components (agency-risk index, semantic telemetry, continuous authorization, FSM conformance, drift detection, graduated containment). It governs deployed-agent execution, we govern the upstream version lifecycle, and the two compose: MI9-style execution-time governance fits inside our Stage~6 (online monitoring), while compatibility checking and gated activation operate at the admission boundary MI9 does not address. Behavioral Contracts~\citep{behavioralcontracts} formalize per-agent run-time obligations as machine-checkable contracts; this is structurally close to our policy compatibility check $\kappa_P$, but their contracts apply within one agent version, whereas $\kappa_P$ compares the policy footprint of a candidate against that of the active version, surfacing upgrade-time policy drift that intra-version contracts do not.

The governed upgrade pipeline shares structural parallels with established software deployment practice. Staged rollout~\citep{stagedrollout}, canary and blue-green patterns~\citep{canary}, and automated rollout with reinforcement learning~\citep{autorollout} incrementally expose a new version while the old one remains active. The DevOps performance literature~\citep{accelerate} establishes deployment frequency, change-failure rate, and mean time to restore as organizational metrics; our pipeline-level metrics (BADR, FAR, UAR, RSR, PVR, SRDR) are the embodied counterpart that evaluate whether a single upgrade decision is admissible. Semantic versioning research~\citep{semver,semsemver} shows that roughly a third of library releases introduce breaking changes despite version labels, motivating automated compatibility validation; progressive delivery and feature flags~\citep{progressivedelivery,featuretoggles} decouple deployment from release with metric-gated promotion, matching our candidate-registration and gated-activation stages. Chaos engineering~\citep{chaosengineering} parallels our fault-injection sandbox but targets infrastructure rather than capability-level behavioral drift.

ML deployment monitoring finds the same lifecycle pressures in non-embodied settings: hidden technical debt catalogues post-training failure modes~\citep{sculley2015techdebt}, the ML Test Score turns them into a production-readiness rubric (the closest SE precedent for our four-dimensional model)~\citep{breck2017mltest}, and deployment surveys treat versioning, rollback, and monitoring as continuous concerns~\citep{paleyes2022challenges,ashmore2021assuring}. At the substrate layer, LITHE hot-swaps control laws into a 1\,kHz loop without interruption~\citep{lithe}; it solves the swap mechanism, while we govern which candidate is admissible to swap in.

Stage by stage, the pipeline has deployment-practice counterparts (flag enrollment, pre-production testing, canary traffic, ramp-up, alerting, automated rollback), but embodied capability upgrade adds requirements absent from web-service and MLOps pipelines, beginning with physical-world safety: two versions must never drive the same actuator. Supplementary Material \S K tabulates the comparison concern by concern, and \Cref{sec:discussion} returns to what can and cannot be borrowed.

\paragraph{Position of this work}
AEROS~\citep{aeros}, capability-centric evolution~\citep{identity}, and runtime governance~\citep{harnessing} together supply an architectural layer between cognition and execution, modular evolvable capabilities, and runtime safety as a systems concern. What none of them supplies is the lifecycle layer between them: how a new capability version enters a running embodied system under governance. That is the gap this paper addresses, and it is not a direct application of DevOps or MLOps practice. The question is no longer only whether a proposed action should be allowed, but whether a newly produced version should be admitted, shadow-tested, activated, monitored, and, if necessary, rolled back.

\section{Problem Formulation}
\label{sec:problem}

\subsection{Background: System Model}
\label{sec:pf-setting}

We consider a long-lived embodied agent system with one persistent agent, a set of versioned capability modules, and a runtime governance layer~\citep{aeros,identity,harnessing}. The system at time $t$ is
\begin{equation}
    \mathcal{R}_t = (\mathcal{A},\; \mathcal{C}_t,\; \policyset_t,\; \Gamma_t),
    \label{eq:system}
\end{equation}
where $\mathcal{A} = (\pi_{\mathcal{A}}, M_{\mathcal{A}}^{\mathrm{id}}, M_{\mathcal{A}}^{\mathrm{ep}})$ is the persistent agent: $\pi_{\mathcal{A}}$ is its fixed decision policy, $M_{\mathcal{A}}^{\mathrm{id}}$ its identity memory (who the agent is: credentials, roles, and self-descriptive state), and $M_{\mathcal{A}}^{\mathrm{ep}}$ its episodic memory (what the agent has experienced: the accumulated execution history it consults at decision time). All three are invariant throughout capability evolution; upgrades change capabilities, not the agent. $\mathcal{C}_t = \{c_1^{(k_1)}, \ldots, c_n^{(k_n)}\}$ is the active capability set where each $c_i^{(k_i)}$ is the currently active version of the $i$-th \ecm{}, $\policyset_t$ is the runtime policy configuration, and $\Gamma_t$ is the governance context (deployment profile, authority state, environment constraints). Upgrade decisions are conditioned on the governing state $G_t = (\policyset_t, \Gamma_t)$; the same candidate may be admissible in a simulation profile but rejected in a human-shared setting~\citep{harnessing}.

During operation, the system may produce a candidate $\candidate_i^{(k+1)}$ through RL, imitation learning, LLM-based synthesis, or manual revision. The key issue is not how the candidate is learned but how it is \emph{admitted}: even a candidate that improves task success may change invocation structure, expand permissions, alter trace distributions, or invalidate recovery assumptions. We therefore distinguish \emph{capability production} from \emph{capability admission} (whether the candidate may enter the active system, under what mode, and with which monitoring conditions).

\subsection{Upgrade Decision Function}
\label{sec:pf-decision}

We formalize capability upgrade admission as a governance decision over the current active version, the candidate version, and the current runtime state. Let
\begin{equation}
    \upgfunc\bigl(\candidate_i^{(k+1)},\, c_i^{(k)},\, \policyset_t,\, \Gamma_t\bigr) \to \bigl\{\texttt{reject},\, \texttt{sandbox},\, \texttt{shadow},\, \texttt{activate},\, \texttt{rollback-review}\bigr\}
\label{eq:upgrade-decision}
\end{equation}
denote the upgrade decision function. Its five verdicts differ in how much trust the candidate has earned. A \texttt{reject} verdict means the candidate is not admissible under current compatibility or policy conditions. A \texttt{sandbox} verdict admits it only to isolated evaluation, where the candidate can be exercised without touching the live execution path. A \texttt{shadow} verdict lets the candidate run in parallel observation mode: it receives the same inputs as the active version but controls nothing. An \texttt{activate} verdict allows the candidate to replace or augment the active version, always under continued runtime monitoring. Finally, \texttt{rollback-review} couples tentative activation to rollback readiness or supervisory review, for contexts where no automatic promotion is acceptable.

This decision function differs from ordinary execution-time admission~\citep{harnessing}, which decides whether a proposed capability invocation may execute now. Here, the decision concerns whether a new version may become part of the executable system itself; we formalize the consequences of this distinction in the next subsection.

\begin{definition}[Governed Capability Evolution]
\label{def:gce}
A long-lived embodied system exhibits \emph{governed capability evolution} if every candidate capability version is admitted into active execution only after explicit evaluation of its compatibility across four dimensions (interface, policy, behavioral, and recovery; see \Cref{sec:compatibility}), and remains subject to monitoring and rollback after activation.
\end{definition}

This separates governed evolution from \emph{na\"ive upgrade} (immediate replacement after training) and \emph{offline-only validation} (one-shot checks with no post-activation monitoring). Upgrade admissibility is context-dependent: the same candidate may be accepted, shadowed, or rejected depending on deployment profile and policy configuration.

\subsection{Failure Modes of Na\"ive Upgrade}
\label{sec:pf-failure}

Suppose the system activates $\candidate_i^{(k+1)}$ without explicit governance. Four classes of failure may follow. A renamed parameter or retyped output silently invalidates existing consumers, so the planner or dispatcher breaks at the call site (\emph{interface breakage}). The new version widens its permission footprint and acts where existing runtime policies cover it insufficiently (\emph{policy breakage}). Behavior regresses beneath a healthy-looking surface, with nominal task success improving while unsafe continuation and anomalous traces increase (\emph{behavioral regression}). Or the upgrade weakens the recovery assumptions the runtime depends on, so when the version eventually fails the system can no longer roll back, fall back, or safely interrupt (\emph{recovery degradation}). These four classes motivate treating capability upgrade as a governed lifecycle event rather than a purely optimization-driven update, and they define the in-taxonomy fault families used in the evaluation (\Cref{sec:experiments}).

\subsection{Design Objective}
\label{sec:pf-objective}

The objective of this paper is not to maximize upgrade acceptance rate at any cost. Instead, we seek an embodied upgrade process that jointly satisfies five properties. Beneficial capability versions should still be deployable (\emph{improvement}), and unsafe or policy-incompatible upgrades should be prevented (\emph{safety}). Upgraded capabilities must remain structurally usable by the current system (\emph{compatibility}). Failed or drifting upgrades should admit rollback or fallback (\emph{recoverability}), and every upgrade decision and post-activation intervention should remain inspectable after the fact (\emph{auditability}).

Formally, the target is not simply $\max\,J(\mathcal{R}_{t+1})$ over system utility $J$, but admitting only candidates that improve utility while preserving the governance envelope over execution and failure handling, extending policy-constrained execution from action-level to capability-level deployment control.

\section{Upgrade Compatibility Model}
\label{sec:compatibility}

\subsection{Overview}
\label{sec:compat-overview}

The central premise of this paper is that a newly produced capability version should not be treated as an immediate replacement for the currently active one. Instead, it should be evaluated as a governed candidate before entering the active capability set. To make this decision analyzable, we introduce an upgrade compatibility model that decomposes capability admission into four complementary dimensions: interface compatibility, policy compatibility, behavioral compatibility, and recovery compatibility.

This decomposition is motivated by the architectural commitments established in prior work. In the single-agent embodied architecture~\citep{aeros}, capabilities are packaged as explicit, installable modules rather than being fused into a monolithic controller. In the capability-evolution view~\citep{identity}, modules are versioned, updated, and gated over time while the agent identity remains fixed. In the runtime-governance view~\citep{harnessing}, execution remains subject to admission, policy checking, monitoring, recovery, and audit. The missing step is to determine whether a new capability version remains compatible with the surrounding system along all of these dimensions before activation.

Let $c_i^{(k)}$ denote the currently active version of capability $i$, and let $\candidate_i^{(k+1)}$ denote a candidate upgraded version. We define the compatibility assessment function
\begin{equation}
    \mathcal{C}\bigl(\candidate_i^{(k+1)},\; c_i^{(k)},\; \policyset_t,\; \Gamma_t\bigr) \;\to\; (\kappa_I,\;\kappa_P,\;\kappa_B,\;\kappa_R),
    \label{eq:compat-function}
\end{equation}
where $\kappa_I$, $\kappa_P$, $\kappa_B$, and $\kappa_R$ denote compatibility outcomes for interface, policy, behavioral, and recovery dimensions, respectively. The overall upgrade decision is then derived from these compatibility outcomes rather than from raw task performance alone.

\subsection{Capability Representation}
\label{sec:compat-representation}

To support upgrade analysis, each capability version is represented by both executable logic and machine-readable operational metadata. This extends the \ecm{} abstraction introduced in prior work~\citep{aeros,identity}, where modules expose interfaces, permissions, risk information, rollback support, and environment-related descriptors.

We represent a capability version $c$ as
\begin{equation}
    c = (I_c,\; O_c,\; \phi_c,\; P_c,\; B_c,\; R_c,\; D_c),
    \label{eq:cap-repr}
\end{equation}
where:
\begin{itemize}[leftmargin=*,nosep]
    \item $I_c$ and $O_c$ are the input and output interface specifications;
    \item $\phi_c$ is the invocation schema and declarative capability descriptor;
    \item $P_c$ is the permission and policy-relevant profile;
    \item $B_c$ is the behavioral signature inferred from execution traces;
    \item $R_c$ is the recovery profile, including rollback and safe-abort assumptions;
    \item $D_c$ is deployment metadata such as version, dependencies, and environment scope.
\end{itemize}

The purpose of this representation is not to fully formalize the internals of every capability implementation, but to expose enough structure for upgrade-time governance. The tuple is the working data structure of the compatibility model: each component is consumed by exactly one of the four dimensions defined next. Interface compatibility $\kappa_I$ compares $(I_c, O_c, \phi_c)$ across versions, policy compatibility $\kappa_P$ examines $P_c$ against the active policy set, behavioral compatibility $\kappa_B$ compares the signatures $B_c$, and recovery compatibility $\kappa_R$ evaluates $R_c$. The deployment metadata $D_c$ carries the environment scope against which a deployment profile is selected; the prototype's activation controller reads the profile rather than $D_c$ itself.

\subsection{Four Compatibility Dimensions}
\label{sec:compat-dimensions}

We define the four dimensions in parallel below. An overview table summarizing the core question, checks, and outcomes for each dimension is provided in Supplementary Material \S K.

\paragraph{Interface compatibility ($\kappa_I$)}
A new version may break the surrounding system even if it improves task performance. We evaluate
\begin{equation}
\begin{split}
\kappa_I &= \textsc{Compat}_I\!\bigl(c_i^{(k)},\, \candidate_i^{(k+1)}\bigr) \\
         &= f_I\!\bigl(\Delta_{\mathrm{sig}},\, \Delta_{\mathrm{schema}},\, \Delta_{\mathrm{pre/post}},\, \Delta_{\mathrm{dep}}\bigr),
\end{split}
\label{eq:kappa-I}
\end{equation}
where the four $\Delta$ terms capture deviation between the candidate's and the active version's interface components $(I_c, O_c, \phi_c)$ of \Cref{eq:cap-repr}: input/output signatures, invocation schema, pre/postconditions, and dependency requirements. A version is \emph{compatible} if the current system can invoke it without modification, \emph{conditionally compatible} if bounded adaptation suffices, and \emph{incompatible} if activation would require planner or runtime changes beyond the upgrade-governance scope.

\paragraph{Policy compatibility ($\kappa_P$)}
Even if callable, a candidate may exceed the current policy envelope~\citep{harnessing}. We evaluate
$\kappa_P = \textsc{Compat}_P(\candidate_i^{(k+1)}, \policyset_t, \Gamma_t)$,
checking the declared permission profile $P_c$ of \Cref{eq:cap-repr} against the active policy set: permissions requested beyond those already granted, and execution parameters outside the tolerated band. The prototype scores both and requires the combined value to clear the profile's $\kappa_P$ threshold. A fuller check would also quantify coverage of reachable execution modes and refuse any uncovered high-risk mode; we leave that to future work (\Cref{sec:threats}).

\paragraph{Behavioral compatibility ($\kappa_B$)}
Structural callability and policy coverage do not guarantee stable runtime behavior. We emphasize that ``behavioral compatibility'' here refers to \emph{trace-level alignment} between a candidate version's execution profile and the system's governance expectations (success rates, timing, anomaly incidence, recovery behavior), not to ``behavioral safety'' in the safe RL sense~\citep{saferl}, which concerns reward-shaping or constrained optimization during policy learning. Our notion operates at the deployment level: a version may be safe in the RL sense yet behaviorally incompatible with the running system's governance assumptions. We compare behavioral signature vectors
\begin{equation}
    B_c = \bigl(\mu_{\mathrm{succ}},\, \mu_{\mathrm{time}},\, \mu_{\mathrm{retry}},\, \mu_{\mathrm{viol}},\, \mu_{\mathrm{anom}},\, \mu_{\mathrm{recover}}\bigr),
\label{eq:behavioral-sig}
\end{equation}
derived from execution traces $\mathcal{T}^{\mathrm{old}}$ and $\mathcal{T}^{\mathrm{new}}$, computing a drift function $\Delta_B = d(B_{\mathrm{old}}, B_{\mathrm{new}})$. The upgrade is behaviorally suspicious if it improves nominal success while sharply worsening safety-critical dimensions. A candidate may be behaviorally incompatible even when its average task success exceeds the active version. This is a key reason why governed evolution cannot be reduced to benchmark maximization.

\paragraph{Recovery compatibility ($\kappa_R$)}
A high-performing candidate may degrade system robustness if the runtime can no longer recover from its failures~\citep{identity,harnessing}. We evaluate
\begin{equation}
\kappa_R = \textsc{Compat}_R\!\bigl(\candidate_i^{(k+1)},\, c_i^{(k)},\, \policyset_t,\, \Gamma_t\bigr),
\label{eq:kappa-R}
\end{equation}
checking the recovery profile $R_c$ of \Cref{eq:cap-repr}: rollback, fallback, safe-abort, and failure-mode recognizability. We define a recovery-readiness score
$\rho(\candidate) = g(r_{\mathrm{rollback}}, r_{\mathrm{fallback}}, r_{\mathrm{abort}}, r_{\mathrm{detect}})$
over the hooks declared in $R_c$ and the recovery component of the candidate's declared signature; low $\rho$ restricts activation mode or requires review. This remains a manifest-level check: a candidate retaining a hook key whose implementation no longer functions can still clear it, a blind spot probed empirically in \Cref{sec:results}. Critically, recoverability is an \emph{admission} criterion, not merely a post-failure concern.

\subsection{Compatibility Composition}
\label{sec:compat-composition}

The four dimensions are separated because upgrade failure emerges in different ways: a candidate may be structurally callable but behaviorally unstable, policy-covered but unrecoverable, or nominally high-performing yet inadmissible under real-robot deployment policy.

We therefore define overall compatibility not as a single scalar but as a governed composition:
\begin{equation}
    \mathcal{C}_{\mathrm{overall}} = \Psi(\kappa_I,\; \kappa_P,\; \kappa_B,\; \kappa_R),
    \label{eq:composite-compat}
\end{equation}
where $\Psi$ is a governance aggregation rule. In our framework, aggregation is conservative:
\begin{itemize}[leftmargin=*,nosep]
    \item any interface incompatibility yields immediate rejection;
    \item policy incompatibility yields rejection or review depending on environment profile;
    \item behavioral incompatibility yields sandbox-only or shadow-only restriction;
    \item recovery fragility yields activation only under rollback-coupled monitoring.
\end{itemize}

This conservative composition reflects the principle that capability activation is a system-level commitment, not a pure performance-optimization step.

\subsection{Compatibility Outcomes and Deployment Modes}
\label{sec:compat-outcomes}

The compatibility model produces one of four deployment-oriented outcomes: \textbf{fully compatible} (eligible for gated activation), \textbf{conditionally compatible} (admissible under bounded conditions such as simulation-only operation, stricter thresholds, shadow-first deployment, or mandatory approval), \textbf{evaluation-only} (not activatable, but informative to exercise in sandbox or shadow), and \textbf{incompatible} (rejected or returned for revision). These outcomes feed the pipeline of \Cref{sec:pipeline}. The model unifies the gating intuitions of earlier capability-evolution and runtime-governance work~\citep{identity,harnessing} into an explicit upgrade-time decision structure, making precise how upgrade admission differs from execution admission.

\section{Governed Upgrade Pipeline}
\label{sec:pipeline}

\subsection{Overview}
\label{sec:pipe-overview}

The compatibility model defines what must be checked before a new version enters the active system; this section operationalizes those checks as a governed lifecycle. A candidate does not move directly from production to activation: it passes through a staged pipeline in which structural compatibility, policy sufficiency, behavioral stability, and recovery readiness are evaluated under increasingly realistic execution conditions.

This pipeline extends runtime governance from execution-time action control to capability-lifecycle control: the runtime mediates not only whether an installed capability may execute now, but whether a newly produced version may join the active capability substrate at all. Upgrade becomes a controlled systems transition rather than a local optimization event.

\Cref{fig:overview} contrasts this path with na\"ive replacement. The seven stages are: (1)~candidate registration, (2)~compatibility validation, (3)~sandbox evaluation, (4)~shadow deployment, (5)~gated activation, (6)~online monitoring and drift handling, and (7)~rollback, demotion, and audit closure, a progression from low-risk evaluation to active deployment in which a candidate may advance, repeat, pause, or terminate at any stage.

\subsection{Pipeline Stages}
\label{sec:pipe-stages}

The seven stages and their roles are summarized in Supplementary Material \S K; below we provide the key formal elements for each stage.

\paragraph{Stage 1: Candidate registration}
A new version $\candidate_i^{(k+1)}$ enters the version registry $\mathcal{V}_i$ with provenance, declared interface changes, permission profile, dependency set, and estimated risk level. Registration marks the lifecycle boundary: a version is managed before it is executable~\citep{identity}.

\paragraph{Stage 2: Pre-activation compatibility validation}
The four compatibility dimensions from \Cref{sec:compatibility} are evaluated:
$(\kappa_I, \kappa_P, \kappa_B, \kappa_R) = \mathcal{C}(\candidate_i^{(k+1)}, c_i^{(k)}, \policyset_t, \Gamma_t)$.
Validation routes the candidate to fail-fast reject, sandbox-only, shadow-eligible, or activation-eligible; interface and policy incompatibility reject immediately.

\paragraph{Stage 3: Sandbox evaluation}
Candidates run in an isolated environment $\mathcal{E}_{\mathrm{sb}}$ under canonical tasks and structured perturbations (noise, delay, tool unavailability), computing the metrics that drive admission:
\begin{equation}
    \mathcal{S}(\candidate_i^{(k+1)}) = \{m_{\mathrm{succ}},\, m_{\mathrm{viol}},\, m_{\mathrm{anom}},\, m_{\mathrm{retry}},\, m_{\mathrm{recover}}\}.
\label{eq:sandbox-metrics}
\end{equation}
Compatibility is necessary, not sufficient: a candidate may pass the checks yet fail sandbox.

\paragraph{Stage 4: Shadow deployment}
The candidate runs in parallel with the active version on the same input stream but controls nothing, evaluating comparative divergence $\Delta_{\mathrm{shadow}}(t) = d(y^{\mathrm{old}}_t, y^{\mathrm{new}}_t)$ for regression discovery, drift discovery, and live-context policy checking, bridging the distributional gap between sandbox isolation and real deployment.

\paragraph{Stage 5: Gated activation}
Only after shadow deployment is a candidate eligible for promotion. Activation proceeds only if
\begin{equation}
    \theta_{\mathrm{act}}(\kappa_I,\kappa_P,\kappa_B,\kappa_R,\mathcal{S},\Delta_{\mathrm{shadow}},\Gamma_t) = 1,
\label{eq:activation-gate}
\end{equation}
and is profile-sensitive: one candidate may be activated in simulation, conditionally activated on a real robot, and rejected in a human-shared setting~\citep{harnessing}.

\paragraph{Stage 6: Online monitoring and drift handling}
An activated upgrade remains provisional; monitoring evaluates the post-activation telemetry $\Omega_t^{\mathrm{upgrade}}$:
\begin{equation}
    M(\Omega_t^{\mathrm{upgrade}}, \policyset_t, \Gamma_t) \to \{\texttt{continue},\, \texttt{restrict},\, \texttt{escalate},\, \texttt{rollback}\},
\label{eq:monitor-decision}
\end{equation}
tracking performance and policy drift, behavioral anomaly, and recovery instability.

\paragraph{Stage 7: Rollback, demotion, and audit closure}
If monitoring detects unacceptable drift, the system restores the prior active version:
$\mathcal{C}_t \leftarrow (\mathcal{C}_t \setminus \{c_i^{(k+1)}\}) \cup \{c_i^{(k)}\}$.
Hard rollback (immediate removal), soft demotion (back to sandbox or shadow), and profile demotion (restriction to lower-risk environments) are distinguished; audit closure records the lifecycle.

\subsection{Pipeline as a Governance State Machine}
\label{sec:pipe-fsm}

\begin{figure*}[pos=t!p]
\centering
\includegraphics[width=0.92\textwidth]{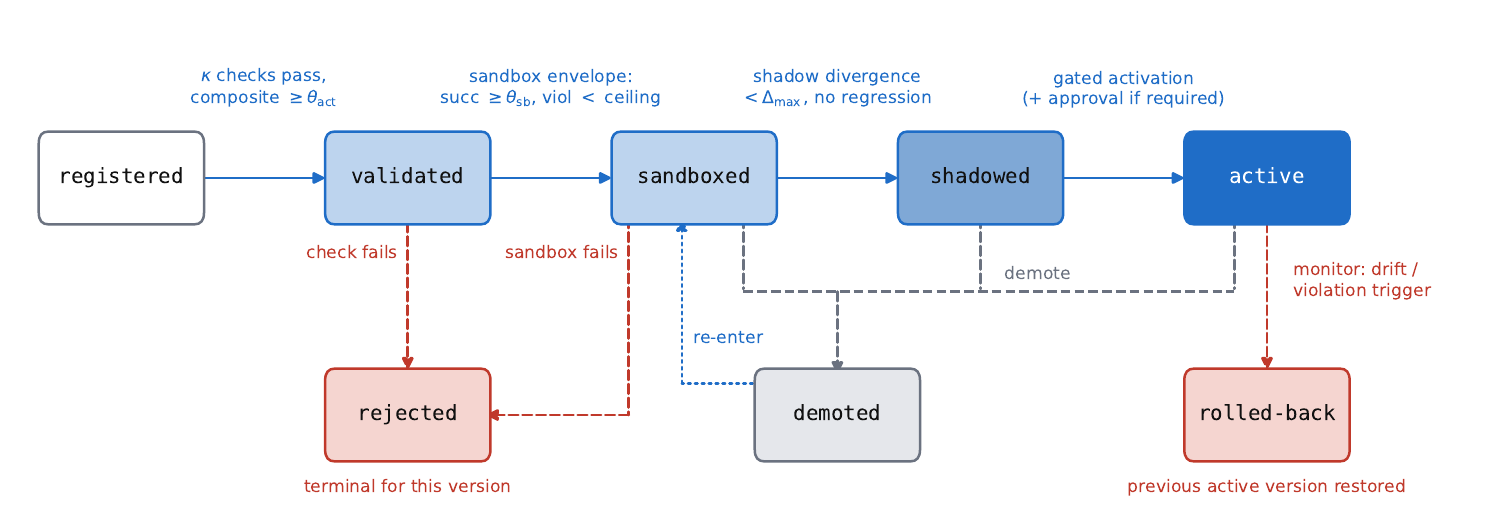}
\caption{Governance state machine over candidate capability versions. Forward transitions (left to right) require passing the stage's admission criterion; downward transitions capture rejection, demotion, and rollback. \texttt{active} remains monitored: drift returns a candidate to \texttt{rolled-back} or \texttt{demoted}, and \texttt{demoted} candidates re-enter at sandbox evaluation, the stage the experiments exercise.}
\label{fig:fsm}
\end{figure*}

The seven stages above are specified as a governance state machine over candidate capability versions. It is a design specification: the prototype exercises its stage components (\Cref{sec:implementation}) rather than an integrated orchestrator (\Cref{sec:threats}). \Cref{fig:fsm} shows the states and their transitions. Each candidate occupies one of eight lifecycle states: \texttt{registered}, \texttt{validated}, \texttt{sandboxed}, \texttt{shadowed}, \texttt{active}, \texttt{demoted}, \texttt{rejected}, and \texttt{rolled-back}. Transitions are driven by compatibility outcomes and runtime evidence rather than by a ``train then replace'' rule. Promotion is gated at every step: all four compatibility dimensions and the composite score must clear their thresholds to reach \texttt{validated}; the sandbox envelope $\theta_{\mathrm{sb}}$ and violation ceiling must hold across canonical, perturbed, and adversarial modes to reach \texttt{sandboxed}; trace divergence from the active version must stay below $\Delta_{\mathrm{max}}$ with no regression category firing to reach \texttt{shadowed}; and approval-requiring profiles add supervisory sign-off before \texttt{active}. Any failed check routes the candidate to \texttt{rejected} or \texttt{demoted}. The state that matters most is \texttt{active}, which is not absorbing: online monitoring returns a live candidate to \texttt{rolled-back} or \texttt{demoted} when windowed violation or anomaly rates cross the monitoring triggers, so governance stays engaged after deployment rather than ending at promotion. \texttt{rejected} and \texttt{rolled-back} are terminal for that version; a \texttt{demoted} candidate re-enters at the stage it was demoted to. Supplementary Material \S A gives the full transition table with the trigger parameters.

\section{Prototype Implementation}
\label{sec:implementation}

\subsection{Base Architecture and Persistent Core}
\label{sec:impl-arch}

The prototype is a reference implementation, not a production stack. It is built on three architectural commitments from prior work: a single persistent embodied agent~\citep{aeros}, modular capability packaging through \ecm{}s~\citep{identity}, and a runtime governance layer that mediates execution under explicit policy constraints~\citep{harnessing}. Upgrade governance is added as a lifecycle layer above existing modular runtimes without modifying the persistent agent.

The reference architecture has five subsystems. The \textbf{Persistent Agent Core} performs goal interpretation, task planning, capability selection, and invocation dispatch through the runtime governance layer rather than executing directly; it remains the unique decision-making subject and is not modified during upgrade, preserving identity invariance~\citep{identity}. The \textbf{\ecm{} Registry and Loader} manages installed capability packages and versions, exposing manifests, interfaces, dependency metadata, and activation state. The \textbf{Runtime Governance Layer} continues to provide capability admission, policy enforcement, runtime watching, recovery coordination, and audit logging~\citep{harnessing}. The \textbf{Upgrade Manager} is the architectural role this paper introduces: it owns candidate registration, compatibility validation, sandbox evaluation, shadow deployment, activation gating, and rollback. The \textbf{Execution Backend} provides the simulator-facing substrate; the architecture remains platform-agnostic.

\begin{figure*}[pos=t!p]
\centering
\includegraphics[width=0.95\textwidth]{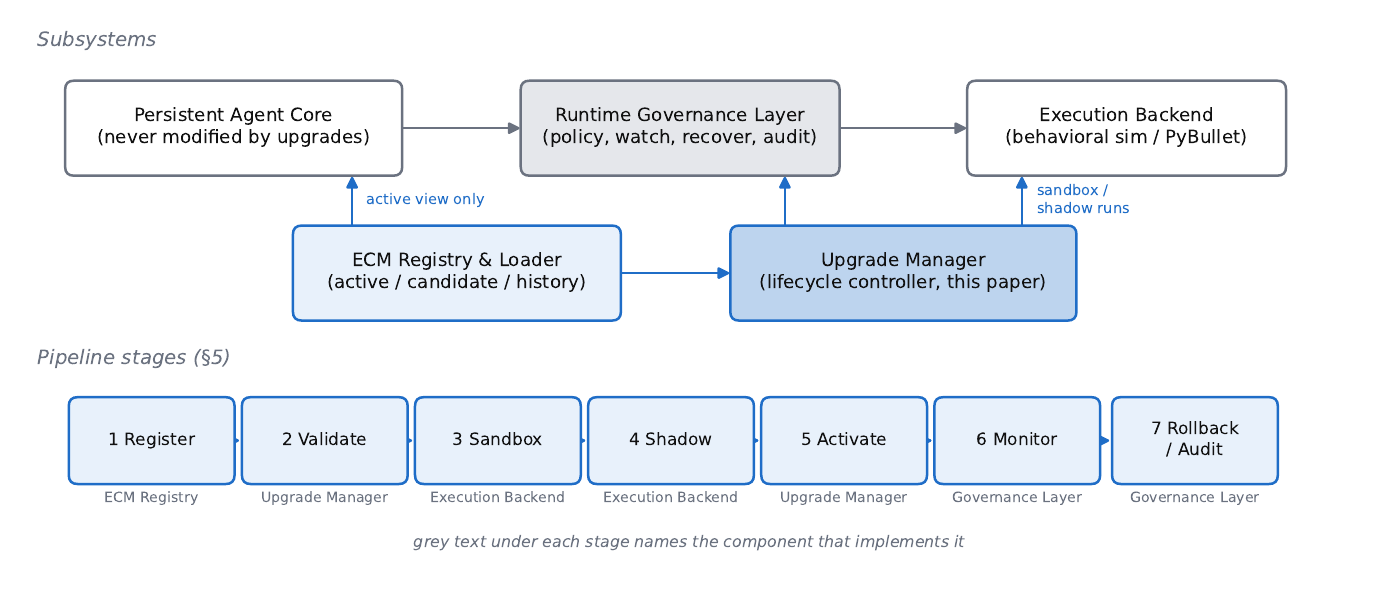}
\caption{Reference architecture and stage-to-component mapping. Top: the five subsystems; the persistent agent core is never modified by upgrades, and candidate versions stay under Upgrade Manager control until gated activation promotes them into the registry's active view. Bottom: the seven formal pipeline stages of \Cref{sec:pipeline}, each labelled with the component that implements it. The evaluated artifact implements the stage components and the registry; the remaining boxes are architectural context.}
\label{fig:arch}
\end{figure*}

Two of these five are inherited from the predecessor runtimes (summarized in Supplementary Material \S J) and are not re-implemented here. What the artifact provides is the Upgrade Manager's stage components (\Cref{sec:impl-compat}) plus a version registry, and the experiment drivers compose those components in stage order; the persistent agent core, the active dispatcher, and a persistent audit store are architectural context rather than evaluated code. \Cref{fig:arch} maps the formal pipeline stages of \Cref{sec:pipeline} onto that architecture: registration and compatibility validation belong to the Upgrade Manager and read registry metadata; sandbox and shadow evaluation execute through dedicated executors on the execution backend; and activation, monitoring, and rollback close the loop through the runtime governance layer, which retains execution-time authority over whatever version is active.

\subsection{ECM Packaging and Upgrade Manager}
\label{sec:impl-registry}

Each capability is packaged as an \ecm{} with a manifest (name and version, input/output interface schema, invocation entry points, dependency declarations, permission profile, environment scope, rollback metadata, optional recovery hooks) and an executable implementation bundle. The registry stores both active and candidate versions:
\begin{equation}
    \mathcal{V}_i = \{c_i^{(0)},\; c_i^{(1)},\; \ldots,\; c_i^{(k)},\; \candidate_i^{(k+1)},\; \ldots\},
    \label{eq:registry}
\end{equation}
Three registry views (\emph{active}, \emph{candidate}, \emph{history}) make lifecycle state a first-class object. In the architecture each entry carries lifecycle state, provenance, validation history, deployment status, and audit links (Supplementary Material \S F); the artifact's records are thinner, holding the module, creation and promotion timestamps, status, and notes, which is what the rollback experiment needs. Candidates are never loaded directly into the active dispatcher path; they stay under Upgrade Manager control until activation is explicitly approved.

The Upgrade Manager is a lifecycle controller. For each candidate $\candidate_i^{(k+1)}$ it maintains a state variable $z(\candidate_i^{(k+1)}) \in$ \{\,\texttt{registered}, \texttt{validated}, \texttt{sandboxed}, \texttt{shadowed}, \texttt{active}, \texttt{demoted}, \texttt{rejected}, \texttt{rolled-back}\,\}, with transitions driven by compatibility results and runtime evidence rather than by a ``latest version wins'' rule. The lifecycle of \Cref{sec:pipeline} is realized as stage components (compatibility checker, sandbox and shadow executors, activation controller, online monitor, rollback controller) that the experiment drivers compose in stage order; a production deployment would wrap them behind one lifecycle-controller interface. The Upgrade Manager sits above the existing execution-time governance structure, giving the runtime two governance boundaries: execution-time (whether an active capability may run now) and upgrade-time (whether a new version may become active at all). Of the state machine's full verdict set, the experimental controller exercises reject, activate, and review-gated activation; demotion and re-entry exist as registry states but are not exercised by the experiments in this paper.

\subsection{Compatibility Checker and Evaluation Executors}
\label{sec:impl-compat}

The Compatibility Checker operationalizes the four-dimensional model from \Cref{sec:compatibility} over declared manifest metadata: an \textbf{Interface Checker} matching the candidate's interface schema field-by-field against the active version's, a \textbf{Policy Checker} comparing permission scope and execution parameters against the current $\policyset_t$ (allow, restricted, approval-required, or reject), a \textbf{Behavioral Checker} comparing declared signatures, and a \textbf{Recovery Checker} scoring the candidate's coverage of the active version's recovery-hook set together with its declared recovery quality. Two scope notes matter for reading the results. The prototype's $\kappa_P$ is profile-independent: it accepts a $\Gamma_t$ argument but averages permission-scope and execution-parameter compatibility without reading the profile, so the profile-dependent behavior in \Cref{sec:res-e5} comes from profile-owned admission thresholds, sandbox ceilings, and approval requirements applied outside the checker. And $\kappa_B$ scores the manifest's declared signature; sandbox and shadow statistics gate the candidate at their own stages rather than feeding back into $\kappa_B$. Supplementary Material \S B gives the checking procedure in full, and \S G the per-template scores on the development set.

The \textbf{Sandbox Executor} runs candidates in isolation with its own invocation wrapper, telemetry channel, and policy envelope, supporting three modes (canonical-task, perturbation, adversarial-governance). Each run emits a structured trace record (capability version, task instance, inputs and outputs, duration, retry count, anomaly flags, policy hits, recovery triggers). The \textbf{Shadow Executor} performs live-context parallel evaluation: both the active and candidate versions receive the same input stream but only the active version controls execution. The Shadow Executor records output divergence, governance-signal divergence, and trace-envelope divergence.

\subsection{Activation, Monitoring, Rollback, and Audit}
\label{sec:impl-activation}

The \textbf{Activation Controller} consumes compatibility outcomes, sandbox and shadow metrics, environment profile, and authority mode through threshold- and rule-based gating; decisions are explicit and inspectable rather than learned end-to-end. The \textbf{Online Upgrade Monitor} reuses the runtime watcher infrastructure with deployment-phase counters for upgraded versions (rolling success, policy warnings, anomalies, rollback triggers); threshold crossings trigger downgrade, restriction, or rollback. The \textbf{Rollback Controller} restores the prior active version, repairs dispatch state, and emits audit events, supporting hard rollback and soft demotion. An \textbf{Audit and Telemetry Store} is specified to hold each candidate's record from provenance to final outcome; the experiment drivers emit the per-candidate decision records the analysis consumes.

The prototype does not yet provide a fully formal capability type system, probabilistic policy verification, multi-robot orchestration, or autonomous policy synthesis. Its role is to show that \Cref{sec:problem,sec:compatibility,sec:pipeline} can be instantiated with a coherent software structure and evaluated experimentally.

\subsection{Computational Overhead}
\label{sec:impl-overhead}

We benchmarked per-stage wall-clock cost on the behavioral backend over twenty timed repetitions after a discarded warm-up, single-threaded on an Apple M1 Max under Python 3.11.15 and NumPy 2.4.6; \texttt{scripts/benchmark\_overhead.py} regenerates the table and ships the raw repetitions in \texttt{data/overhead\_benchmark.json}, and Supplementary Material \S K gives the breakdown with sample sizes. Compatibility checking was timed over the full 135-candidate pool, the three execution-bearing stages over a fixed 15-candidate subset, and monitoring over one 50-task rollback window. Per candidate: compatibility check $0.023 \pm 0.001$\,ms, sandbox $1.60 \pm 0.06$\,ms, shadow $1.70 \pm 0.03$\,ms, gated activation $0.003 \pm 0.000$\,ms; online monitoring costs $0.198 \pm 0.002$\,ms per window. Screening all 135 candidates takes $6.3 \pm 0.9$\,ms and one E2 seed $93.2 \pm 1.1$\,ms. Unlike the experimental results, these figures are hardware- and load-dependent and do not reproduce byte-identically; the shipped record fixes the run the table reports.

Pre-activation governance therefore costs about 3.3\,ms per candidate, of which the governance logic proper (compatibility checking and activation gating) is under 1\%; sandbox and shadow execution account for the rest. Tripling the batch from 45 to 135 candidates leaves the per-candidate compatibility cost flat within noise (0.0234 versus 0.0232\,ms), so cost tracks candidates evaluated rather than stages traversed. The physics track is far slower per candidate because inverse-kinematics solving and physics stepping dominate; we did not benchmark it, so these figures describe the behavioral backend only.

\section{Experimental Setup}
\label{sec:experiments}

\subsection{Experimental Objective}
\label{sec:exp-objective}

The evaluation tests whether success-signal deployment strategies admit unsafe capability upgrades that governed upgrade prevents, and what the prevention costs. The experiments therefore measure not only whether an upgraded capability performs better than an older one, but whether upgraded capabilities can be safely admitted, monitored, and, when necessary, rolled back within a long-lived embodied system.

\subsection{Evaluation Platform}
\label{sec:exp-platform}

We evaluate the prototype in a two-track simulation setup. The primary track is a behavioral simulation of an embodied manipulation stack: each capability version carries an operating profile (success, timing, retry, violation and anomaly propensities, recovery quality), and the environment turns that profile into stochastic execution traces over parameterized grasp, align, and place task instances with randomized object properties, arm states, and tolerances. This track is deliberately abstract: it runs the complete governance pipeline at millisecond cost per candidate, which is what makes 45-candidate pools per family and seed, five strategies, and 15-seed replication practical. The secondary track is a PyBullet physics backend, where a Kuka IIWA arm executes the same task suite through inverse-kinematics control in deterministic DIRECT mode, with violations and anomalies surfaced from joint states and contacts; it replicates the headline strategy comparison to check that decision-layer findings are not artifacts of the behavioral abstraction. Both tracks implement the same environment interface, run the identical upgrade pipeline, and inject stochastic perturbations into object positions, observation noise, action latency, and execution disturbances.

\subsection{Task Suite}
\label{sec:exp-tasks}

We use a small but structured embodied task suite composed of modular manipulation tasks, chosen to satisfy three properties: tasks must require reusable capability modules rather than monolithic end-to-end control; tasks must allow measurable improvement across versions; tasks must expose failure modes that depend on policy and on recovery behavior.

The task suite includes three task families. \textbf{Grasp} requires the robot to approach and grasp an object under pose variation. \textbf{Align} requires aligning an object with a target slot, tray, or marker under positional uncertainty. \textbf{Place} requires moving and placing the object into a designated region while respecting workspace and motion constraints.

In some experiments, we also compose these into a pick-align-place sequence to test whether upgrade effects generalize beyond isolated capability execution and remain governable over longer task horizons. Each task instance is randomized over object placement, initial arm state, target position, and environmental perturbations.

\subsection{Capability Modules Under Evolution}
\label{sec:exp-ecms}

We instantiate the capability framework using versioned \ecm{}s, each corresponding to a reusable functional skill family: \ecm{}-Grasp, \ecm{}-Align, and \ecm{}-Place. For each capability family $i$, the system begins with an active baseline version $c_i^{(0)}$. Through simulated refinement rounds, we generate a sequence of candidate upgraded versions:
\begin{equation}
    c_i^{(0)} \to \candidate_i^{(1)} \to \candidate_i^{(2)} \to \cdots
    \label{eq:version-seq}
\end{equation}
Candidate versions are created through parameter refinement, control-logic modification, or configuration-level adjustment. The exact update mechanism is not the primary variable; what matters is that each new version can potentially change task performance as well as interface assumptions, runtime behavior, permission requirements, and recovery characteristics.

\subsection{Candidate Generation}
\label{sec:exp-candidates}

How evaluation candidates are constructed determines how much the evaluation can claim: hand-picking faulty candidates that match the checked fault classes would make it self-fulfilling. We therefore generate the evaluation pool from \emph{parameterized fault operators} with seeded sampling: an operator family defines how a candidate may deviate from the active version, and concrete candidates are drawn by sampling the operator's parameter space (which schema fields drift, which permissions are added, how far behavioral parameters move, which recovery hooks degrade) together with a continuous severity level. The pool is therefore determined by the experiment seed rather than by author choice, it differs across seeds, and it includes marginal, threshold-straddling instances by construction. Supplementary Material \S C specifies the operators and their parameter ranges.

Benign candidates come from a contract-preserving improvement operator (bounded gains in success, timing, retries, and recovery quality; interfaces and permissions unchanged), testing whether governed upgrade stays practically useful rather than merely conservative. Faulty candidates are sampled from four operator families that instantiate the failure classes of \Cref{sec:pf-failure}. The \emph{interface drift} operator applies one to three sampled schema edits (rename, retype, removal, or an added required input; additive-only drift is deliberately included because it is invisible to subset-based interface checking). The \emph{permission expansion} operator requests one to three extra permissions drawn from a risk-weighted universe, with unsafe behavior scaled by the sampled risk. The \emph{behavioral regression} operator produces nominally attractive candidates whose violation and anomaly propensities grow with severity while trajectories become more aggressive. The \emph{recovery degradation} operator weakens recovery hooks in one of two modes: removing the hook from the manifest, or retaining the declared key while disabling the hook, a manifest-level blind spot probed deliberately.

Beyond these four in-taxonomy families, the generator adds \emph{out-of-taxonomy probes}: latency inflation (slower execution with an otherwise nominal profile), rare intermittent faults (violation probabilities sitting below screening statistics), and context-dependent faults (nominal on two task types, unsafe on the third, so aggregate statistics dilute the signal). None of the four compatibility checks models these classes; they measure where the pipeline's coverage ends rather than confirming what it already checks.

Per capability family and seed, the pool contains 12 benign, 24 in-taxonomy faulty (6 per operator family), and 9 out-of-taxonomy candidates. Manifests are truthful: declared signatures derive from actual operating profiles, so detection depends on a fault's statistical footprint crossing a threshold, not on manifest deception. The hand-built templates are retained solely as the threshold development set (\Cref{sec:exp-protocol}); they never enter evaluation pools.

\subsection{Deployment Strategies Compared}
\label{sec:exp-baselines}

We compare five deployment strategies, chosen so that the comparison spans the two extremes and the established middle ground of deployment practice. \textbf{Static} keeps the initial versions fixed: trivially safe, forgoing every improvement. \textbf{Na\"ive upgrade} lets each drawn candidate replace the active version immediately, with no screening or staging.

Two \emph{practice-adapted} baselines port established deployment patterns to capability upgrades, monitoring the aggregate health signal at the core of those patterns; production deployments often add further service metrics, represented below by the policy-aware variant (\Cref{sec:res-e2}). \textbf{Canary} serves a cohort of 20\% of the round's tasks with the candidate while the active version processes the same conditions as a mirrored control; the candidate is promoted to the remaining traffic only if its cohort success rate is not worse than the control's by more than five percentage points, and is aborted otherwise. \textbf{Blue-green} cuts all traffic over to the candidate immediately, keeps the previous version warm, and reverts the remainder of the round if the candidate's success rate over a 20-task monitoring window falls more than five points below the mirrored incumbent. Neither adaptation performs compatibility screening or inspects policy, anomaly, or recovery signals. That is the point of the comparison, not an omission.

\textbf{Governed upgrade (ours)} sends every drawn candidate through the pre-activation admission stack: registration, four-dimensional compatibility checking, sandbox evaluation, shadow deployment, and gated activation. E2 measures admission-decision quality; the post-activation stages are measured separately in E4. All five strategies face the same per-round candidate from the same pool under identically initialized noise, so per-round comparisons are paired.

\subsection{Deployment Profiles}
\label{sec:exp-profiles}

Three deployment profiles differ in three governance parameters: the four compatibility thresholds together with the composite activation gate, the sandbox violation ceiling, and whether activation requires approval. \emph{Simulation} is most permissive ($\kappa_I/\kappa_P/\kappa_B/\kappa_R \geq 0.90/0.85/0.80/0.75$, $\theta_{\mathrm{act}} \geq 0.75$, ceiling 0.20, no approval), \emph{strict runtime} tightens all of them ($0.95/0.92/0.88/0.85$, $\theta_{\mathrm{act}} \geq 0.85$, ceiling 0.15), and \emph{human-shared} is strictest ($0.98/0.95/0.92/0.90$, $\theta_{\mathrm{act}} \geq 0.90$, ceiling 0.10) and requires approval, whose simulated reviewer decides from observable telemetry only, never from ground-truth labels. They test whether one candidate is treated differently across governance contexts; Supplementary Material \S D lists the profile parameters the pipeline consumes.

\subsection{Evaluation Protocol}
\label{sec:exp-protocol}

Each upgrade round is an independent admission trial of one $c^{(k)}\!\to\!\candidate^{(k+1)}$ decision against the persistent baseline: the strategy decides the drawn candidate's fate and serves the round with the resulting version, then the incumbent resets to baseline. Chained evolution, where an accepted candidate becomes the next incumbent, is outside this protocol (\Cref{sec:threats}). Round~0 executes the initial system on the task suite and gives every strategy an identical baseline. At each upgrade round $r \in \{1,\ldots,5\}$: (1)~one candidate version is drawn from the generated pool of the round's capability family, the same candidate for every strategy; (2)~each strategy decides the candidate's fate by its own rule (ignore, activate immediately, canary-promote or abort, blue-green keep or revert, or run the governed pipeline); (3)~the resulting system executes 150 randomized task instances; (4)~all runtime telemetry, policy events, anomalies, and recovery outcomes are recorded. Within a round, every strategy starts from an identically reseeded noise stream: comparisons are paired at the seed-and-round level, though streams drift once outcomes differ (no per-task common-random-number claim).

We run 15 random seeds; each re-samples the candidate pool, the draw sequence, the task instances, and the noise streams, so replicates vary in all factors at once. Reported metrics are means with standard deviations over seeds; final-round contrasts use two-sided paired Wilcoxon tests with Holm--Bonferroni correction across the comparison family. All compatibility thresholds, activation gates, and monitoring triggers are fixed a priori at documented operating points. The four thresholds and the composite gate are additionally documented against the development set's score separation (every benign template clears every threshold, every faulty template fails at least one), which the artifact's calibration script re-derives; sandbox, shadow, and monitoring parameters are configuration choices reported in the supplement. All are disjoint from the generated evaluation pools (\Cref{sec:exp-candidates}); no threshold tuning is performed on evaluation data.

\subsection{Metrics}
\label{sec:exp-metrics}

Because this paper studies governable upgrade rather than pure task optimization, we use both performance metrics and governance metrics. The six primary metrics are defined here; auxiliary metrics (execution time, retry count, recovery latency) are in Supplementary Material \S E.

Let $F$ be the set of faulty candidates evaluated, $B$ the set of benign candidates, $A(\cdot)$ the indicator that a candidate was activated into the live execution path, $T$ the set of executed task instances, and $R$ the set of triggered rollback attempts. The six metrics are:
\begin{equation}
\begin{aligned}
\mathrm{BADR} &= \tfrac{1}{|F|}\textstyle\sum_{f \in F} \bigl(1 - A(f)\bigr), &
\mathrm{FAR} &= \tfrac{1}{|B|}\textstyle\sum_{b \in B} \bigl(1 - A(b)\bigr), \\
\mathrm{UAR} &= \tfrac{\sum_{x \in F} A(x)}{\sum_{x \in F \cup B} A(x)}, &
\mathrm{SR} &= \tfrac{1}{|T|}\textstyle\sum_{\tau \in T} \mathrm{succ}(\tau), \\
\mathrm{RSR} &= \tfrac{1}{|R|}\textstyle\sum_{r \in R} \mathrm{restored}(r), &
\mathrm{PVR} &= \tfrac{1}{|T|}\textstyle\sum_{\tau \in T} \mathbb{1}[\mathrm{viol}(\tau) > 0],
\end{aligned}
\label{eq:metrics}
\end{equation}
where $\mathrm{succ}(\tau)$ indicates task success, $\mathrm{restored}(r)$ indicates that rollback restored the prior active version with a verified post-rollback state, and $\mathrm{viol}(\tau)$ counts policy violations in trace $\tau$. BADR (bad-upgrade detection rate) is recall over faulty candidates; FAR (false-alarm rate) is the fraction of benign candidates wrongly kept out of activation; UAR (unsafe activation rate) is the faulty share of activated candidates, the fraction of what actually reached live execution that should not have. In per-round reporting (E2), UAR is the indicator that the round's activated candidate was faulty. BADR and FAR are computed over in-taxonomy candidates; out-of-taxonomy probes are reported separately, being outside the model's design scope. Together, these metrics reveal whether a strategy achieves improvement by quietly sacrificing governability.

\subsection{Experiment Groups}
\label{sec:exp-groups}

We organize the evaluation into five experiment groups:

\textbf{E1: Upgrade Screening.}
Tests whether static compatibility screening alone detects faulty candidates before any dynamic evaluation. Main outcomes: bad-upgrade detection rate, false reject rate, per-family interception rates.

\textbf{E2: Performance--Safety Tradeoff.}
Compares Static, Na\"ive, Canary, Blue-green, and Governed Upgrade over repeated upgrade rounds. Main outcomes: success rate, violation rate, and unsafe activation rate per round, plus the faulty-exposure fraction of the practice-adapted baselines.

\textbf{E3: Shadow Deployment Effectiveness.}
Measures whether shadow execution identifies regressions that would otherwise appear only after deployment. Main outcomes: shadow regression detection rate, avoided bad activations.

\textbf{E4: Rollback Reliability Under Drift.}
Activates upgraded capabilities and then injects runtime drift (observation corruption, delay, shifted object distributions). Main outcomes: rollback success rate, recovery latency, unsafe continuation after rollback triggers.

\textbf{E5: Cross-Profile Upgrade Governance.}
Tests whether the same candidate version is correctly admitted, restricted, or rejected across different deployment profiles. Main outcome: profile-sensitive admissibility consistency.

\section{Results}
\label{sec:results}

We evaluate whether governed capability evolution improves deployment safety and at what cost, comparing five strategies: Static, Na\"ive Upgrade, the practice-adapted Canary and Blue-green baselines, and Governed Upgrade (Ours). The short version of the findings: the success-signal baselines capture nearly all of the nominal performance benefit and admit faulty versions in 61.3--74.7\% of upgrade rounds while doing so; governance admits no in-taxonomy fault, and one out-of-taxonomy probe, at a 4.6-point success cost; and out-of-taxonomy probes locate the boundary of that guarantee. All numbers below are from 15 seeds with generated candidate pools; the artifact regenerates the experiment result files with documented commands, byte-reproducibly across processes, and the bundled analysis, ablation, calibration, and sensitivity scripts regenerate every reported statistic from them.

\subsection{E1: Upgrade Screening}
\label{sec:res-e1}

E1 evaluates whether the compatibility model detects problematic upgrades before any dynamic evaluation, across the four in-taxonomy fault-operator families. Screening is the first filter of the pipeline, not its whole defense, so the numbers here bound what static checks alone contribute; \Cref{sec:res-ablation} composes them with the dynamic stages.

\begin{table*}[t]
\centering
\caption{Compatibility-screening results over the generated pools (15 seeds; 72 in-taxonomy faulty and 36 benign candidates per seed; 405 out-of-taxonomy probes pooled). Rejection rates are pooled across seeds with Wilson 95\% confidence intervals.}
\label{tab:screening}
\smallskip
\begin{tabular*}{\linewidth}{@{\extracolsep{\fill}}lccc@{}}
\toprule
\textbf{Fault-operator family} & \textbf{Rejected / total} & \textbf{Rate} & \textbf{95\% CI} \\
\midrule
Interface drift        & 243/270 & 90.0\% & [85.8, 93.0] \\
Permission expansion   & 189/270 & 70.0\% & [64.3, 75.2] \\
Recovery degradation   & 168/270 & 62.2\% & [56.3, 67.8] \\
Behavioral regression  & 51/270  & 18.9\% & [14.7, 24.0] \\
\midrule
Benign (false alarms)  & 0/540   & 0.0\%  & [0.0, 0.7]   \\
Out-of-taxonomy probes & 0/405   & 0.0\%  & [0.0, 0.9]   \\
\bottomrule
\end{tabular*}
\end{table*}

Compatibility screening alone rejects 60.3\%\,$\pm$\,4.7\% of in-taxonomy faulty candidates (BADR) at zero false-alarm rate over benign candidates. \Cref{tab:screening} breaks the number down by operator family, and the decomposition is informative. Structurally explicit faults are caught at high rates: interface drift at 90.0\% (the misses are additive-only schema edits, invisible to subset-based checking by construction), permission expansion at 70.0\% (single-permission requests sit inside the tolerance band), and recovery degradation at 62.2\% (the disable-in-place mode retains declared hook keys and evades manifest-level coverage checking). Behavioral regressions, in contrast, are caught at only 18.9\% by static screening: their manifests are truthful but nominally attractive, and small signature drifts stay inside the $\kappa_B$ tolerance. Every out-of-taxonomy probe passes screening, as expected by construction, since none of the four checks models those classes. Screening is an effective, intentionally incomplete structural filter; the dynamic stages carry the rest, quantified next.

\subsection{E2: Performance--Safety Tradeoff}
\label{sec:res-e2}

E2 compares the five deployment strategies over repeated upgrade rounds. The question has two sides: whether governance prevents the unsafe activations that success-signal deployment lets through, and what that protection costs in task performance.

\begin{table*}[t]
\centering
\caption{Deployment outcomes across upgrade rounds (mean $\pm$ std over 15 seeds; round-0 baseline plus five upgrade rounds). Mean UAR averages the per-round unsafe-activation indicator over the five upgrade rounds. Evaluation exposure is the mean share of upgrade-round traffic served by a faulty candidate during the strategy's own evaluation phase, before its promote or revert rule fired; Static and Na\"ive have no evaluation phase, and Governed evaluates in sandbox and shadow rather than on served traffic.}
\label{tab:deployment}
\begin{tabular*}{\linewidth}{@{\extracolsep{\fill}}lccccc@{}}
\toprule
\textbf{Metric} & \textbf{Static} & \textbf{Na\"ive} & \textbf{Canary} & \textbf{Blue-green} & \textbf{\textcolor{paperblue}{Governed}} \\
\midrule
Final SR (\%)$\uparrow$    & 66.2{\scriptsize$\pm$3.4} & 73.3{\scriptsize$\pm$5.2}  & 73.2{\scriptsize$\pm$5.3}  & 73.3{\scriptsize$\pm$5.2}  & 68.7{\scriptsize$\pm$5.6} \\
Final UAR (\%)$\downarrow$ & 0.0{\scriptsize$\pm$0.0}  & 73.3{\scriptsize$\pm$44.2} & 66.7{\scriptsize$\pm$47.1} & 73.3{\scriptsize$\pm$44.2} & 0.0{\scriptsize$\pm$0.0} \\
Mean UAR (\%)$\downarrow$  & 0.0                        & 74.7                        & 61.3                        & 64.0                        & 1.3 \\
Final PVR (\%)$\downarrow$ & 11.1{\scriptsize$\pm$3.3} & 22.6{\scriptsize$\pm$14.1} & 22.3{\scriptsize$\pm$14.3} & 22.6{\scriptsize$\pm$14.1} & 11.1{\scriptsize$\pm$3.2} \\
Evaluation exposure (\%)$\downarrow$ & --- & --- & 14.9 & 10.0 & 0.0 \\
\bottomrule
\end{tabular*}
\end{table*}

\begin{figure*}[pos=t!p]
\centering
\includegraphics[width=0.98\textwidth]{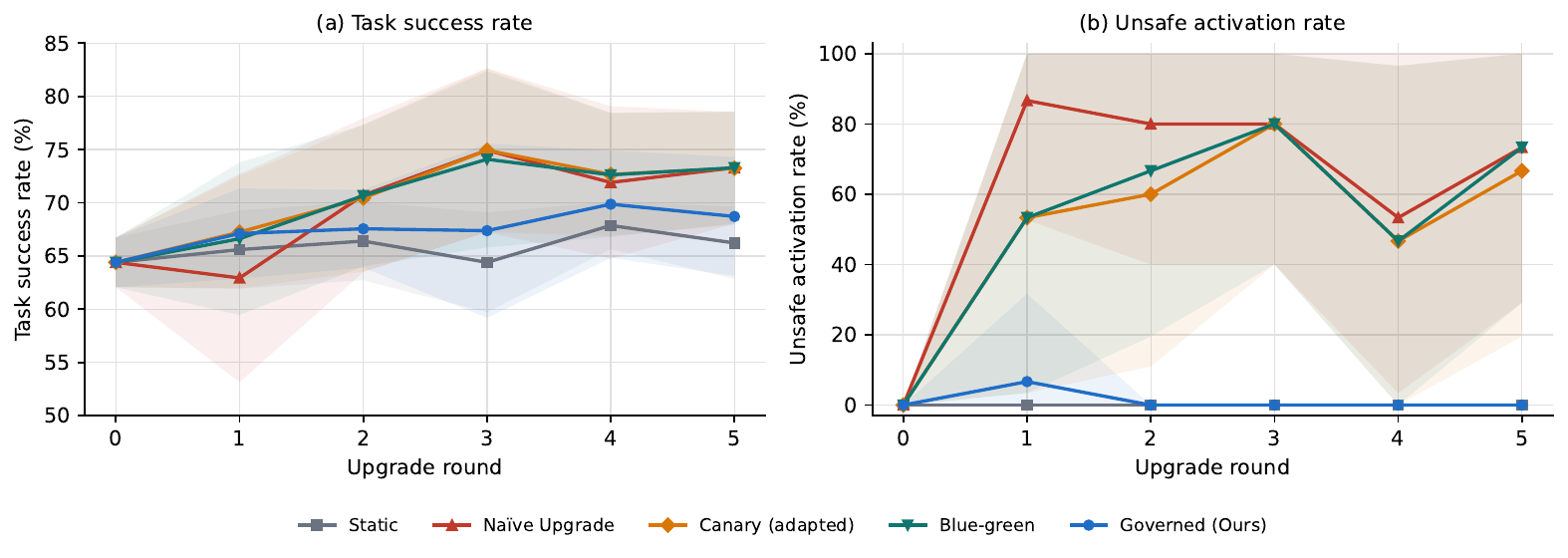}
\caption{Performance and deployment safety over upgrade rounds for the five strategies (mean, with $\pm 1$ standard deviation bands over 15 seeds, clipped to the feasible 0--100\% range; every strategy faces the same per-round candidate and noise stream, so round-0 values coincide by construction).
(a)~Task success rate. Na\"ive, Canary, and Blue-green track each other closely because all three promote most nominally attractive candidates; Governed tracks Static more closely because it rejects most candidates, benign ones included.
(b)~Unsafe activation rate. The success-signal strategies admit faulty candidates in 47--87\% of rounds (per-round means across the three strategies), Canary and Blue-green sitting below Na\"ive in most rounds but far above zero, whereas Governed admits exactly one candidate, an out-of-taxonomy probe, across all 75 upgrade rounds (mean UAR 1.3\%) and no in-taxonomy fault.}
\label{fig:upgrade-curves}
\end{figure*}

\Cref{tab:deployment} and \Cref{fig:upgrade-curves} carry the paper's central comparison. The three success-signal strategies end within a tenth of a point of each other on final-round success (73.2--73.3\%), and they pay for it: Na\"ive activates a faulty candidate in 74.7\% of upgrade rounds, Blue-green in 64.0\%, and Canary in 61.3\%. The practice-adapted baselines help at the margin, aborting or reverting candidates whose success rate visibly collapses (Canary aborts 10 of its 56 faulty draws), with evaluation-phase faulty exposure held to 10--15\% of round traffic. What they cannot see is the failure mode this problem is about: a candidate whose nominal success is equal or better while its policy violations, anomalies, or recovery quality degrade. Such candidates sail through a success-only gate, so Canary and Blue-green admit a faulty version in roughly three of five upgrade rounds and their final-round PVR roughly doubles relative to Static. Governed Upgrade admits one \emph{faulty} candidate in 75 upgrade rounds, an out-of-taxonomy probe, and no in-taxonomy fault, while ending 2.5 points above Static on success.

\paragraph{Statistical analysis}
Two-sided paired Wilcoxon signed-rank tests on per-seed final-round values ($n{=}15$), Holm--Bonferroni corrected across the eight-contrast family, confirm the pattern for the three success-signal contrasts. On UAR, Governed differs from Na\"ive and from Blue-green ($-73.3$\,pp each, $p_{\mathrm{holm}}{=}0.007$, $d_z{=}1.66$) and from Canary ($-66.7$\,pp, $p_{\mathrm{holm}}{=}0.009$, $d_z{=}1.41$). On SR, Governed pays a significant cost against all three ($-4.5$ to $-4.6$\,pp, $p_{\mathrm{holm}}{=}0.024$, $|d_z| \approx 0.95$). The two Static contrasts behave as expected: final-round UAR is identical (both zero) and the SR difference ($+2.5$\,pp) does not reach significance ($p_{\mathrm{holm}}{=}0.22$). The six contrasts also hold on per-seed means over all five rounds (SR $-2.6$ to $-3.6$\,pp, UAR $-60.0$ to $-73.3$\,pp; Holm--Bonferroni corrected within that six-contrast family, $p_{\mathrm{holm}} \le 0.016$): the pattern is not a final-round artifact. Per-seed UAR differences are binary with many ties, so we lean on absolute rates rather than effect-size labels.

\paragraph{The cost of governance, honestly stated}
We report the 4.6-point success gap against Na\"ive as a real cost, not a statistical artifact. Its source is visible in the pipeline telemetry: of the 19 benign candidates the governed pipeline drew across all seeds and rounds, 13 were activated, and all 6 rejections happened at the sandbox gate, whose pass envelope includes adversarial-stress modes. Under the deliberately faulty-heavy pool (33 faulty or out-of-taxonomy versus 12 benign per family), aggressive rejection loses little; the same conservatism prices in foregone upgrades. A benign-heavy sensitivity re-run (24 benign, 8 faulty, 3 probes per family; same seeds and protocol) bounds this cost from the other side: Governed rises to 71.2\%\,$\pm$\,6.5 final-round success (+5.0 points over Static, within 3.2 points of Na\"ive) while admitting no faulty candidate at all, whereas Canary and Blue-green still activate faulty candidates in 21--23\% of upgrade rounds even in that friendlier pool. A policy-aware canary variant probes the other direction: given the runtime's violation telemetry (promotion also requires cohort violation rate within five points of the control's), its unsafe-round share drops from 61.3\% to 21.3\%, with the residue concentrated in the classes whose signal a 30-task cohort cannot carry: latency inflation, recovery degradation, and rare intermittent faults. One added signal narrows the gap without closing it. The safety gap is structural to bounded-window signal gating; the performance cost of governance shrinks as the candidate pool improves and is tunable through the sandbox envelope.

\paragraph{Physics-track replication}
The PyBullet track repeats the five-strategy comparison on all 15 seeds, with success emerging from inverse-kinematics execution rather than an operating profile; candidate draws are identical across tracks, so the comparison is like for like. The ordering transfers: Na\"ive admits faulty candidates in 74.7\% of rounds, Blue-green in 54.7\%, Canary in 49.3\%, Governed in 4.0\%. The boundary result transfers only partially: of the three governed admissions in 75 physics-track upgrade rounds, one is an out-of-taxonomy rare-intermittent probe, and two are low-severity in-taxonomy faults (a recovery degradation and a single-permission expansion) whose footprints stay inside the gates under the noisier physics telemetry; each admission's fault category is recorded in the shipped physics results, so the attribution is auditable directly. Absolute success levels are lower and compressed (final-round SR 57.1--61.2\%) because IK execution dominates variance. The transferable conclusion is the ordering, not the absolute zero: success-signal gates promote half to three quarters of faulty upgrades in either track, the governed pipeline admits a few percent, and dynamic-gate fidelity tracks telemetry fidelity, which is why activation remains provisional and monitored (\Cref{sec:res-e4}).

\subsection{E3: Shadow Deployment Effectiveness}
\label{sec:res-e3}

\begin{figure*}[pos=t!p]
\centering
\includegraphics[width=0.9\textwidth]{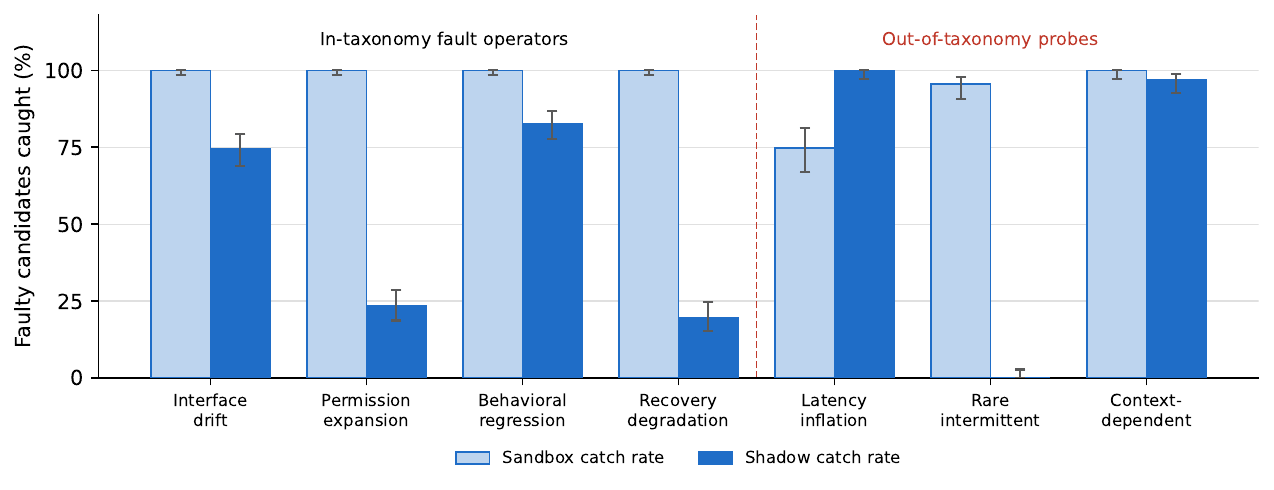}
\caption{Sandbox versus shadow catch rates per fault-operator family (pooled over 15 seeds; error bars are Wilson 95\% intervals). In-taxonomy families are shown left of the divider, out-of-taxonomy probes right. Sandbox evaluation absorbs the statistical bulk across families; shadow deployment's unique contribution concentrates in latency inflation, where candidates that pass the sandbox envelope diverge visibly from the active version under live timing.}
\label{fig:shadow-regressions}
\end{figure*}

\Cref{fig:shadow-regressions} reports, for every fault-operator family, the fraction of faulty candidates caught by sandbox evaluation and by shadow deployment (Wilson 95\% intervals over 15 seeds). Two findings replace the picture the smaller hand-built pool suggested. First, sandbox evaluation is stronger than its reputation: with three evaluation modes including adversarial stress, it catches every candidate from all four in-taxonomy fault families, all context-dependent faults, and 95.6\% of rare intermittent faults. Second, shadow deployment's incremental value is real but concentrated: over the full pool, 2.3\% of faulty candidates (34/1485, CI [1.6, 3.2]) are caught by shadow alone, and every one of those cases is a latency-inflation probe, where 25.2\% (CI [18.6, 33.1]) evade the sandbox's success-and-violation envelope while their live trace-envelope divergence gives them away in shadow. Shadow also remains the only stage able to detect live-context divergence, defined as it is by comparison against the live active version. The residual escapes sit where the design predicts: 4.4\% of rare intermittent faults pass both stages because their statistical footprint stays below both envelopes; those are the candidates that online monitoring and rollback exist to absorb. Defense in depth here is not a slogan but a measured division of labor: screening intercepts structural faults, sandbox absorbs the statistical bulk, shadow closes the latency and live-context corner, and the post-activation stages carry what remains.

\subsection{E4: Rollback Reliability Under Drift}
\label{sec:res-e4}

E4 is a forced-activation stress test of the rollback mechanism: one faulty candidate per fault family is activated directly, deliberately bypassing admission, runtime drift is injected, and once monitoring triggers, the rollback controller must restore the prior active version. Success requires the previous version to be restored and to pass a verification probe under the injected drift; restoration is registry- and dispatch-level, not deep state repair.

The rollback mechanism restores verified operation (a restore-and-verify criterion) in 72.4\%\,$\pm$\,5.4\% of 720 pooled attempts (521 restored; Wilson 95\% CI [69.0, 75.5]; per-drift-type breakdown in Supplementary Material \S K). Actuator delay is most survivable (81.1\%\,$\pm$\,9.4), sensor noise and distribution shift follow (76.7\%\,$\pm$\,8.7, 69.4\%\,$\pm$\,11.7), and combined drift is hardest (62.2\%\,$\pm$\,10.5): verification must pass under all three drifts at once. Failed restorations are close to uniform across the four fault categories (25.0--31.7\% per class), and candidates whose manifests retain declared-but-disabled hooks (the disable-in-place mode) fail slightly more often than the rest (30.6\% versus 27.3\%), a difference too small to carry causal weight at this sample size. The binding constraint is drift severity at verification time, not the candidate that caused the rollback. Even benign-looking candidates can fail after activation, so governance must persist past the activation boundary.

\subsection{E5: Cross-Profile Upgrade Governance}
\label{sec:res-e5}

E5 evaluates whether upgrade admissibility changes appropriately across deployment profiles.

Across the full pipeline, the three profiles trace a monotone safety--throughput frontier (per-profile breakdown in Supplementary Material \S K). Every profile judges the same pinned dynamic evidence per candidate, so the differences below are attributable to the governance parameters alone. The simulation profile detects 97.3\%\,$\pm$\,1.5 of in-taxonomy faults and activates 83.9\% of benign candidates, with faulty versions (probes included) making up 12.9\%\,$\pm$\,4.4 of its activations; the strict runtime profile reaches in-taxonomy BADR 99.6\%\,$\pm$\,0.8 (72.2\% benign activation, faulty share 6.6\%); and the human-shared profile closes the gap entirely, in-taxonomy BADR 100\% and zero faulty activations, at the price of activating only 7.6\% of benign candidates. Per-candidate records make the boundary explicit: in the most permissive profile, 9.6\% of out-of-taxonomy probes reach activation versus 2.7\% of in-taxonomy faults, and under the human-shared profile neither class reaches activation. The human-shared reviewer decides from observable sandbox and shadow telemetry only: its zero-leak result is earned by an extremely conservative evidence bar, not oracle knowledge of candidate labels. The same candidate is admitted in simulation, restricted under strict runtime, and held for review in human-shared contexts; an upgrade cannot be judged in the abstract. The frontier realizes the configured strictness ordering: E5 verifies the implementation delivers it, not that it is emergent.

\subsection{Ablation of the Upgrade Governance Pipeline}
\label{sec:res-ablation}

The ablation isolates the contribution of each pipeline stage. Unlike the execution-time component ablation in prior runtime-governance work~\citep{harnessing}, here we target the upgrade-lifecycle controls (sandbox evaluation, shadow deployment, online monitoring, rollback, and admission with recovery checks): which lifecycle stages are responsible for preventing unsafe admission, detecting regression, and preserving rollback-ready deployment?

\paragraph{Ablated variants and methodology}
We evaluate the pre-activation screening composition exactly, with no analytical approximation: for every candidate in every generated pool (2{,}025 over 15 seeds; BADR and FAR computed over the 1{,}620 in-taxonomy candidates) we record whether it passes the compatibility check, the compatibility check with the recovery dimension removed, sandbox evaluation, and shadow evaluation. A variant's outcome for a candidate is then computed directly, since a candidate is activated under a variant iff it passes every gate that variant includes. Variants: \textbf{Full} (compatibility, sandbox, shadow), \textbf{$-$Shadow}, \textbf{$-$Sandbox}, \textbf{$-$RecoveryCompat} (recovery dimension removed from the compatibility check), \textbf{CompatOnly}, and \textbf{Na\"ive} (no gates). The post-activation stages are ablated by their own experiment: E4 measures rollback and monitoring directly, and removing them simply forfeits those measured recoveries. BADR and FAR are computed over in-taxonomy candidates; the OOT column reports the fraction of out-of-taxonomy probes that reach activation.

\begin{table*}[t]
\centering
\caption{Screening-composition ablation, computed exactly per candidate over the generated pools (mean $\pm$ std over 15 seeds). A candidate is activated under a variant iff it passes every included gate. OOT act.\,=\,out-of-taxonomy probes reaching activation. $\uparrow$\,=\,higher is better; $\downarrow$\,=\,lower is better.}
\label{tab:ablation}
\begin{tabular*}{\linewidth}{@{\extracolsep{\fill}}lcccc@{}}
\toprule
\textbf{Variant} & \textbf{BADR (\%)}$\uparrow$ & \textbf{FAR (\%)}$\downarrow$ & \textbf{UAR (\%)}$\downarrow$ & \textbf{OOT act.\ (\%)}$\downarrow$ \\
\midrule
Full              & 100.0{\scriptsize$\pm$0.0} & 30.6{\scriptsize$\pm$6.3} & 0.0{\scriptsize$\pm$0.0}  & 1.5{\scriptsize$\pm$2.3} \\
$-$Shadow         & 100.0{\scriptsize$\pm$0.0} & 28.0{\scriptsize$\pm$5.2} & 0.0{\scriptsize$\pm$0.0}  & 9.9{\scriptsize$\pm$5.7} \\
$-$Sandbox        & 78.5{\scriptsize$\pm$5.0} & 3.0{\scriptsize$\pm$2.6}  & 30.3{\scriptsize$\pm$5.0} & 34.3{\scriptsize$\pm$1.6} \\
$-$RecoveryCompat & 100.0{\scriptsize$\pm$0.0} & 30.6{\scriptsize$\pm$6.3} & 0.0{\scriptsize$\pm$0.0}  & 1.5{\scriptsize$\pm$2.3} \\
CompatOnly        & 60.3{\scriptsize$\pm$4.7} & 0.0{\scriptsize$\pm$0.0}  & 44.1{\scriptsize$\pm$2.9} & 100.0{\scriptsize$\pm$0.0} \\
\midrule
Na\"ive           & 0.0{\scriptsize$\pm$0.0}  & 0.0{\scriptsize$\pm$0.0}  & 66.7{\scriptsize$\pm$0.0} & 100.0{\scriptsize$\pm$0.0} \\
\bottomrule
\end{tabular*}
\end{table*}

The composition pattern in \Cref{tab:ablation} is sharper than a uniform ``every stage helps'' story. \textbf{Sandbox evaluation is the load-bearing dynamic filter}: removing it collapses in-taxonomy detection from 100\% to 78.5\%, raises the faulty share of activations from zero to 30.3\%, and lets 34.3\% of out-of-taxonomy probes through; it is also where all benign rejections originate (FAR drops from 30.6\% to 3.0\% without it), so the same gate produces both the safety and the conservatism measured in E2. Sweeping its pass envelope (Supplementary Material \S I) separates the two: over $\theta_{\mathrm{sb}} \in [0.55, 0.70]$ unsafe activation stays $\leq$1.3\% and detection $\geq$98.7\%, while false alarms on drawn benign candidates double (22.7\% to 46.2\%). The envelope is a conservatism dial, not a safety dial. \textbf{Shadow deployment's contribution is targeted}: on in-taxonomy faults it is fully redundant with sandbox at these thresholds, but removing it multiplies out-of-taxonomy leakage more than sixfold (1.5\% to 9.9\%), consistent with E3's finding that shadow's unique catches concentrate in latency-type faults. \textbf{Removing the recovery dimension from screening changes nothing at the admission boundary}: at these thresholds, candidates with degraded recovery are intercepted by the behavioral signature's recovery component or by the dynamic stages. We report this null result deliberately rather than claiming a per-stage contribution the data does not show; E4's per-attempt records likewise show no restoration-failure gap large enough to attribute to hook state (30.6\% versus 27.3\%), so we treat the recovery dimension's value as unproven at both boundaries in this harness rather than assert a contribution the data does not show. \textbf{CompatOnly} bounds what static analysis alone buys: 60.3\% detection at zero false alarms, with every dynamic-signature fault class leaking through. Upgrade-governance quality emerges from the composition of a structural filter, a statistical filter, and a live-divergence filter, each covering a different failure geometry.

\label{sec:res-threats}

\section{Discussion}
\label{sec:discussion}

\paragraph{Why governance matters at upgrade time}
Capability evolution should not be understood only as a learning problem. Once capability versions change continuously, the key systems problem shifts from production to deployment: a new version is a change to the executable substrate that may affect interfaces, policy envelopes, behavioral stability, and recovery simultaneously. Execution-time governance decides whether an installed capability may run now; upgrade-time governance decides whether a new version may become part of what runs at all. The gap is concrete: each action of an upgraded grasp capability can pass the runtime policy guard while the version as a whole raises anomaly incidence from $\mu_{\mathrm{anom}}{=}0.05$ to $0.18$ and removes a pre-condition the rollback procedure relied on. Only upgrade-time governance, comparing behavioral signatures and recovery profiles before activation, flags such a candidate. The behavioral-regression operator produces exactly this pattern by construction. Empirically, five of the six such candidates drawn in E2 were promoted by all three success-signal strategies; none passed the governed pipeline.

Four-way compatibility is richer than API compatibility: a callable upgrade may still be policy-incompatible, behaviorally unstable, or unrecoverable.

\paragraph{Where shadow and rollback fit DevOps}
Production tooling for canary release, automated rollback, and traffic splitting~\citep{stagedrollout,autorollout} can implement parts of Stages~1, 5, and 7, but three barriers limit direct adoption: compatibility here is semantic rather than syntactic (schemas, preconditions, and trace comparison in place of endpoints and status codes); sandbox and shadow evaluation need a physics simulator or safety-constrained wrapper rather than a container sidecar; and deployment profiles are environment-specific, a distinction web-service canary frameworks do not carry. Supplementary Material \S K maps the concerns side by side. The governed pipeline is therefore complementary to DevOps tooling: registry, rollback, and audit infrastructure can be borrowed; the compatibility model and evaluation stages are embodied-specific. E2 measures the difference directly: the canary adaptation admitted a faulty version in 61.3\% of upgrade rounds, the governed pipeline once in 75 (an out-of-taxonomy probe).

\paragraph{Implications and end-to-end policies}
End-to-end policies (RT-2~\citep{rt2}, Octo~\citep{openxembodiment}) do not invalidate the framework: a model-version update is itself a deployment candidate, with $\kappa_I$ checking action and observation spaces, $\kappa_P$ checking the declared permission envelope (coverage of reachable behavior would require analysis the prototype does not implement), $\kappa_B$ comparing trace statistics, and $\kappa_R$ verifying that the runtime can still recover from the new model's failure modes. Modular skill composition remains common practice~\citep{saycan}, so governed evolution stays relevant at several granularities. The governance layer is itself not infallible; its own failure modes (incomplete assessment, distribution gap, misconfiguration, monitor blind spots, rollback unavailability) are analyzed in Supplementary Material \S H. A governed lifecycle does not eliminate upgrade risk; it makes risk inspectable, revisable, and recoverable.

\section{Threats to Validity}
\label{sec:threats}

We discuss threats under the four standard categories~\citep{wohlin2012experimentation}.

\paragraph{Internal validity}
All strategies within a round face the same drawn candidate and identically initialized noise streams, so cross-strategy contrasts are paired at the seed-and-round level. E2's governed arm exercises the pre-activation stack only; conclusions about the post-activation stages rest on E4. Thresholds are calibrated on the hand-built development set and never tuned on evaluation data. The screening-composition ablation is computed exactly per candidate, with no analytical approximation. The remaining internal-validity risk sits in the fault operators themselves: severity-to-behavior mappings are author-specified functions; sampling makes instance selection author-independent, but the mapping shapes decide how hard the marginal cases are. Faults harvested from real reinforcement-learning artifacts rather than designed operators are the natural next step.

\paragraph{External validity}
The empirical validation is simulation-only, an important limitation for a paper about deployment safety. The primary track is deliberately a behavioral abstraction: capability versions are operating profiles, not controllers, which is what makes seeded fault-operator pools and full-pipeline replication cheap, and is also a real distance from physical deployment. The PyBullet physics track narrows that distance for the headline comparison, with success emerging from inverse-kinematics execution rather than from a profile, but actuator wear, communication jitter, contact-rich recovery, and human co-presence still manifest differently on hardware, and the sim-to-real gap is precisely the distribution gap shadow deployment is designed to catch. Expected real-robot challenges include timing compatibility (latency-induced drift, possibly an explicit fifth dimension) and recovery feasibility under real contact constraints. Rollback is also treated throughout as a benign-operator mechanism; recent work shows that checkpoint-restore rollback is itself an attack surface, enabling replay of one-time authorizations and re-execution of compensated actions~\citep{acrfence2026}, and our rollback-reliability results do not cover adversarial rollback. A small-scale real-robot pilot is the highest-priority next step. Scope is also limited in four further senses: each upgrade round is an independent admission trial against the baseline incumbent, so chained multi-version evolution (compounding upgrades on a drifted incumbent) is not exercised; the task suite (grasp, align, place) does not exhaust embodied capabilities; the evaluation is single-system rather than fleet-scale, leaving cross-robot rollout, version coordination, and federated rollback unaddressed; and the human-authority model is simplified to an evidence-gated approval rule, with no supervisory workflows, delayed approval, or organizational policy.

\paragraph{Construct validity}
The central construct-validity question for any fault-injection evaluation is circularity: whether the injected faults are, by construction, precisely the ones the pipeline detects. Three design choices bound this risk without eliminating it: operator-space sampling removes instance selection from author control; thresholds are fixed against a disjoint development set; and out-of-taxonomy probes measure the pipeline against fault classes none of its checks models, with the honest result that screening passes all of them (0/405), 1.5\% reach activation through the full pre-activation stack at base thresholds (9.6\% under the most permissive deployment profile), and exactly one probe reached live activation in E2's 75 governed upgrade rounds. What remains author-chosen is the operator families themselves and their severity-to-behavior mappings, which set how hard the marginal instances are; fault classes we did not think to model are by definition absent. Three further scope limits belong here. The evaluation drives the pipeline's stage components through separate experiment drivers rather than one integrated controller: E1--E3 and E5 cover the pre-activation stages, E4 covers monitoring and rollback from a forced activation, and the demotion and re-entry transitions of \Cref{fig:fsm} are specified but not exercised. The recovery dimension has no demonstrated incremental value at either boundary in this harness, as \Cref{sec:res-ablation} reports. And the baseline result establishes that bounded-window aggregate-success gating fails against these operator families; richer production analogues (longer windows, sequential tests, confidence-bound or anomaly-aware gates) are untested, so the broader insufficiency of service-level signals remains a design argument, not an experimental universal. The compatibility model is structured but not fully formal: behavioral and recovery compatibility rely on thresholds and telemetry comparison rather than temporal logic or proof obligations. This is a deliberate choice for a first systems paper and leaves formal upgrade-policy languages as future work. The \ecm{} abstraction relies on manifests and structured metadata rather than a fully general capability type system, so interface compatibility is partly schema-driven and partly inferred. The six-dimensional behavioral signature $B_c$ (\Cref{eq:behavioral-sig}) is an aggregate proxy that misses high-order temporal structure, rare-event statistics, and trajectory shape; richer or learned signatures are a natural extension.

\paragraph{Conclusion validity}
All final-round contrasts are paired Wilcoxon tests with Holm--Bonferroni correction over the eight-contrast family. The six contrasts against the three success-signal strategies survive correction ($|d_z|$ 0.94--1.66); the two Static contrasts are an identical-zero UAR comparison and a non-significant SR difference, as the design predicts. Per-seed UAR values are binary with heavy ties, so we emphasize absolute rates over effect-size labels for those contrasts. Per-category screening and detection rates carry Wilson 95\% intervals over pooled candidates. The main conclusion-validity caveat is the opposite of underpower: with paired noise streams and 15 seeds, small absolute differences can reach significance, so we emphasize effect sizes and absolute rates over $p$-values throughout.

\section{Conclusion}
\label{sec:conclusion}

This paper introduced governed capability evolution, a lifecycle framework for admitting new capability versions into long-lived embodied systems under runtime governance. The core idea is simple: a newly produced capability version is not an immediate replacement but a governed candidate. We operationalized it through a four-dimensional compatibility model (interface, policy, behavioral, recovery) organized into a staged pipeline from candidate registration through gated activation to monitoring and rollback.

The prototype and experiments show what this discipline buys and what it costs. Practice-adapted canary and blue-green deployment, gating on aggregate success, admit faulty versions in 61.3\% and 64.0\% of upgrade rounds; the governed pipeline admits no in-taxonomy fault and one out-of-taxonomy probe in 75 behavioral-track rounds, at a 4.6-point success cost whose origin (conservative sandbox gating of benign candidates) is measured rather than assumed, and which shrinks as the candidate pool improves. Out-of-taxonomy probes locate the boundary of the guarantee, and rollback restores verified safe operation in 72.4\% of post-activation drift scenarios.

More broadly, the contribution is a design principle as much as a specific pipeline: capabilities must be deployable under governance, not merely learnable. Future work includes formal upgrade-policy languages for declaratively specified admission criteria, fleet-scale rollout over a shared governed registry, real-robot validation, and meta-governance of the pipeline itself, keeping governance parameters sound as agent and environment evolve. Pseudocode, fault-operator definitions, policy profiles, and auxiliary metrics are in the Supplementary Material.



\section*{Statements and Declarations}

\noindent\textbf{Funding.} No funding was received to assist with the preparation of this manuscript.

\noindent\textbf{Competing interests.} The authors have no competing interests to declare that are relevant to the content of this article.

\noindent\textbf{Ethics approval and consent to participate.} Not applicable.

\noindent\textbf{Consent for publication.} All authors consent to publication.

\noindent\textbf{Data availability.} The two-track testbed (behavioral and PyBullet backends), candidate generator, configurations, analysis scripts, and all generated results are available at \url{https://github.com/s20sc/governed-capability-evolution}.

\noindent\textbf{Materials availability.} Not applicable.

\noindent\textbf{Code availability.} The reference implementation of the pipeline, compatibility checkers, and evaluation runners is available at \url{https://github.com/s20sc/governed-capability-evolution}.

\noindent\textbf{Author contributions.} X.Q. conceived the governed capability evolution framework, designed the upgrade pipeline, implemented the prototype, and drafted the manuscript. S.L. contributed to the compatibility checker design and evaluation protocol. J.S. and Z.B. contributed to manuscript revision. C.Y. and Z.L. supervised the research and revised the manuscript. All authors reviewed and approved the final manuscript.


\bibliographystyle{cas-model2-names}
\bibliography{references}

\clearpage
\includepdf[pages=-]{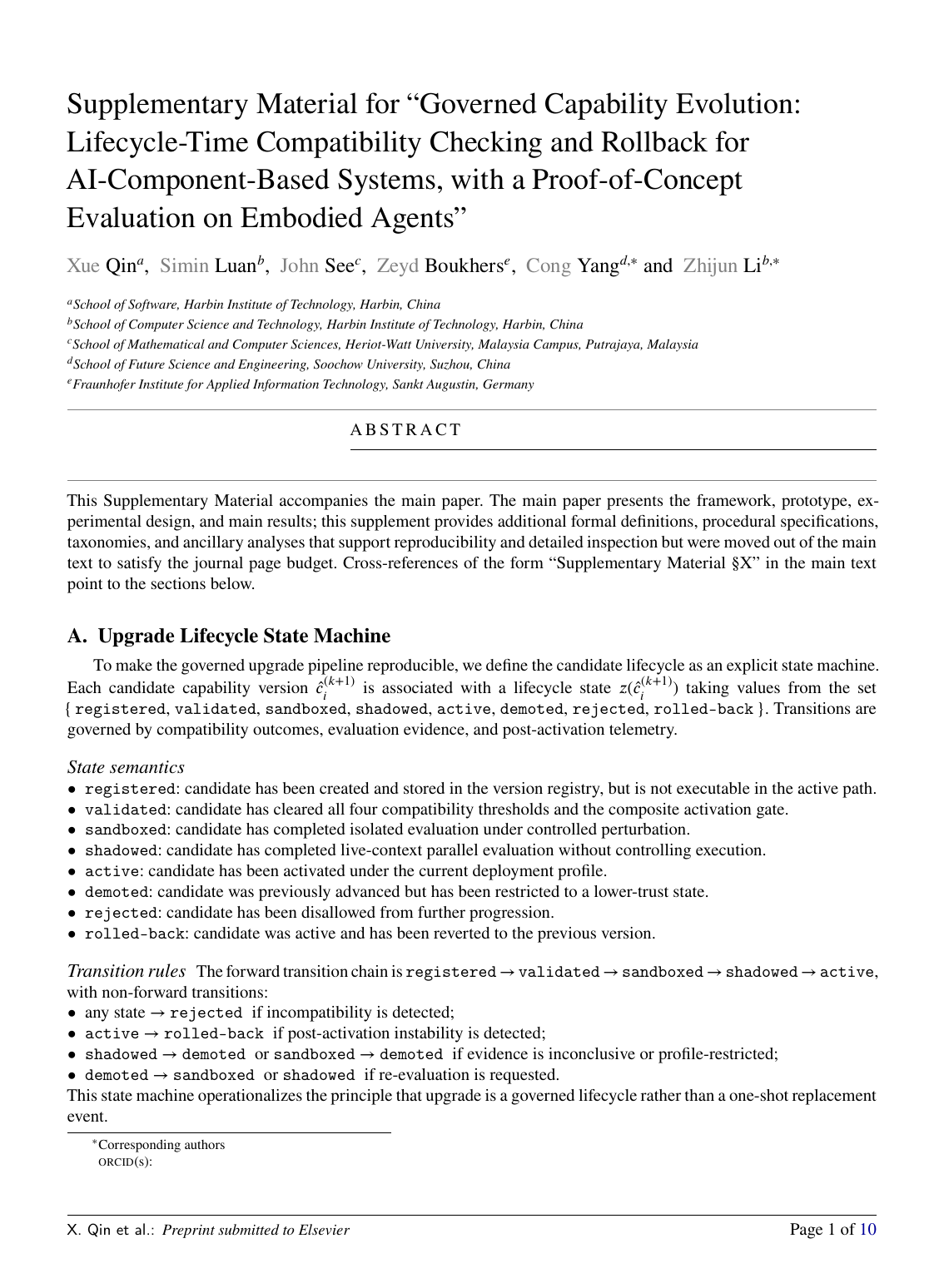}

\end{document}